\documentclass[sigconf, nonacm]{acmart}

\usepackage[utf8]{inputenc} %
\usepackage[T1]{fontenc}    %
\usepackage{hyperref}       %
\usepackage{url}            %
\usepackage{booktabs}       %
\usepackage{amsfonts}       %
\usepackage{nicefrac}       %
\usepackage{microtype}      %
\usepackage{xcolor}         %
\usepackage{balance}

\usepackage{common}
\usepackage{mymacro}
\allowdisplaybreaks

\usepackage[createShortEnv, conf={end, restate, text link section}]{proof-at-the-end} %
\pratendSetGlobal{text link={The \hyperref[proof:prAtEnd\pratendcountercurrent]{proof} is provided in Appendix A.}}

\newtheorem{remark}{Remark}
\newtheorem{example}{Example}

\newtheorem{theorem}{Theorem}
\newtheorem{lemma}[theorem]{Lemma}

\crefname{algocf}{Algorithm}{Algorithms}

\newcommand\vldbdoi{XX.XX/XXX.XX}
\newcommand\vldbpages{XXX-XXX}
\newcommand\vldbvolume{14}
\newcommand\vldbissue{1}
\newcommand\vldbyear{2020}
\newcommand\vldbauthors{\authors}
\newcommand\vldbtitle{\shorttitle} 
\newcommand\vldbavailabilityurl{https://github.com/Guangyi-Zhang/knn-banzhaf-code}
\newcommand\vldbpagestyle{plain}

\begin{document}
\title[Efficient Banzhaf-Based Data Valuation for $k$-Nearest Neighbors Classification]{Efficient Banzhaf-Based Data Valuation for\\$k$-Nearest Neighbors Classification}

\author{Guangyi Zhang}
\authornote{Corresponding author.}
\affiliation{%
  \institution{Shenzhen Technology University}
  \city{Shenzhen}
  \state{China}
}
\email{zhangguangyi@sztu.edu.cn}

\author{Lutz Oettershagen}
\affiliation{%
  \institution{University of Liverpool}
  \city{Liverpool}
  \country{United Kingdom}
}
\email{lutz.oettershagen@liverpool.ac.uk}

\author{Lixu Wang}
\affiliation{%
  \institution{Nanyang Technological University}
  \country{Singapore}
}
\email{wanglixu4334@gmail.com}

\author{Aristides Gionis}
\affiliation{%
  \institution{KTH Royal Institute of Technology \\ Digital Futures}
  \city{Stockholm}
  \country{Sweden}
}
\email{argioni@kth.se}

\begin{abstract}
Data valuation, the task of quantifying the contribution of individual data points to model performance, has emerged as a fundamental challenge in machine learning. 
Game-theoretic approaches, such as the Banzhaf value, offer principled frameworks for fair data valuation; however, they suffer from exponential computational complexity. 
We address this challenge by developing efficient algorithms specifically tailored for computing Banzhaf values in $k$-nearest neighbor (\knn) classifiers.
We first establish the theoretical hardness of the problem by proving that it is \sharpp-hard. 
Despite this intractability, we exploit the locality properties of \knn classifiers to develop practical exact algorithms. 
Our main contribution is a dynamic programming framework that achieves significant computational improvements: 
we present a pseudo-polynomial algorithm with $\bigO(Wkn^2)$ time complexity for weighted \knn classifiers, where $W$ is the maximum sum of top-$k$ weights, and a specialized algorithm for unweighted \knn that achieves $\bigO(nk^2)$ time complexity, that is, linear in the number of data points.
We also offer efficient Monte Carlo estimation methods.
Extensive experiments on real-world datasets demonstrate the practical efficiency of our approach and its effectiveness in data valuation applications.
\end{abstract}

\maketitle
\setcounter{page}{1}

\pagestyle{\vldbpagestyle}
\begingroup\small\noindent\raggedright\textbf{PVLDB Reference Format:}\\
\vldbauthors. \vldbtitle. PVLDB, \vldbvolume(\vldbissue): \vldbpages, \vldbyear.\\
\href{https://doi.org/\vldbdoi}{doi:\vldbdoi}
\endgroup
\begingroup
\renewcommand\thefootnote{}\footnote{\noindent
This work is licensed under the Creative Commons BY-NC-ND 4.0 International License. Visit \url{https://creativecommons.org/licenses/by-nc-nd/4.0/} to view a copy of this license. For any use beyond those covered by this license, obtain permission by emailing \href{mailto:info@vldb.org}{info@vldb.org}. Copyright is held by the owner/author(s). Publication rights licensed to the VLDB Endowment. \\
\raggedright Proceedings of the VLDB Endowment, Vol. \vldbvolume, No. \vldbissue\ %
ISSN 2150-8097. \\
\href{https://doi.org/\vldbdoi}{doi:\vldbdoi} \\
}\addtocounter{footnote}{-1}\endgroup

\ifdefempty{\vldbavailabilityurl}{}{
\vspace{.3cm}
\begingroup\small\noindent\raggedright\textbf{PVLDB Artifact Availability:}\\
The source code, data, and/or other artifacts have been made available at \url{\vldbavailabilityurl}.
\endgroup
}

\section{Introduction}
\label{sec:introduction}
Data valuation has emerged as a fundamental problem in machine learning, addressing the critical question of how to quantify the contribution of individual data points to model performance~\cite{ghorbani2019data,sim2022data,jiang2023opendataval}.
As machine learning systems increasingly rely on large datasets assembled from diverse sources, understanding the relative importance of different data points becomes essential for various applications,
such as enabling practitioners to make informed decisions regarding data acquisition, 
identifying low-quality or redundant data, and ensuring fair compensation for data providers 
in collaborative-learning scenarios.

Cooperative game theory~\citep{elkind2016cooperative} provides a principled framework for addressing data-valuation tasks. 
By modeling the data valuation problem as a cooperative game where data points are players and the model performance represents the value function, we can leverage established solution concepts from cooperative game theory to assign fair and meaningful values to individual data points. 
Among the various solution concepts, the Shapley value~\citep{shapley1953value} and the Banzhaf value~\citep{banzhaf1965weighted} have gained particular attention due to their desirable axiomatic properties. 
In particular, the Banzhaf value has been adopted as a measure of voting power in the analysis of voting in the Council of the European Union~\citep{varela2012negotiating}.

Both values are based on the idea of quantifying the \emph{marginal contribution} of a particular data point to the performance of the model.
Arguably, the simplest way that leverages the same idea to quantify the value of a data point $z$ in a dataset $D$ is the \emph{leave-one-out} (LOO) value, which measures the difference in the model performance before and after the data point is removed from the dataset,
that is,
$
\text{LOO}(z) = v(D) - v(D \setminus \{z\}),
$
where $v: 2^N \to \mathbb{R}$ is the value function that represents the model performance, for example, the accuracy of a model.
However, the LOO value is not informative in the presence of data redundancy~\citep{lee2022deduplicating}.
Besides, the impact of a single data point given the remaining large amount of data is often negligible~\citep{bousquet2002stability,hardt2016train}.

The Shapley and Banzhaf values further develop the idea by aggregating the marginal contributions of a data point to all possible subsets of the dataset.
For example, the Banzhaf value of a data point $z$ is defined precisely as the average marginal contribution of $z$ to all possible subsets $S \subseteq D \setminus \{z\}$ of the dataset, that is,
\[
\bval(z) = \frac{1}{2^{n-1}} \sum_{S \subseteq D \setminus \{z\}} [v(S \cup \{z\}) - v(S)].
\]
The Shapley value is also defined similarly, but with a different weighting scheme over those subsets.
Therefore, these two values give a more comprehensive view of the contribution of a data point to the model performance.

Critically, the Shapley value has been mathematically proved to be the \emph{unique solution} that satisfies the axioms of \emph{efficiency}, \emph{symmetry}, \emph{null player}, and \emph{additivity}~\citep{shapley1953value}, while the Banzhaf value satisfies the same axioms except efficiency.
These axioms are valuable properties that make both Shapley and Banzhaf values natural choices for fair and reliable data valuation.
Among the axioms, efficiency requires that the value of the entire dataset is equal to the sum of the values of all data points, i.e.,
\[
    v(D) - v(\emptyset) = \sum_{z \in D} \sval(z),
\]
where $\sval(z)$ is the Shapley value of data point $z$.
\rrevise{The raw Banzhaf values do not satisfy the efficiency axiom, but this can be remedied by normalizing the raw Banzhaf values, a common post-processing step used in practice~\citep{van1998axiomatizations}.}
\rrevise{It is worth mentioning that though intuitively appealing, efficiency is not essential in many ranking-based applications, including dataset curation and data cleaning, where the focus is on the relative order of data-point values rather than their exact magnitudes.}
\rrevise{For these applications, Banzhaf can be a practical choice: it often yields sparser values and can be less affected by duplicate or noisy points.}
\rrevise{More broadly, the Banzhaf values can be more robust in certain scenarios~\citep{banzhaf1965weighted,wang2023data,li2023robust,karczmarz2022improved}.}

\begin{figure*}[t]
    \centering
    
    \subcaptionbox{Two selected test points (triangle and square) in a 2D dataset.\label{fig:values:dataset}}[0.23\textwidth]{
        {\includegraphics[width=0.23\textwidth]{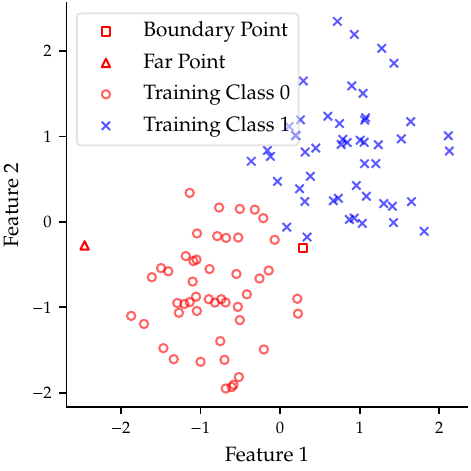}}
    }
    \quad
    \subcaptionbox{The values of the data points with respect to a boundary test point.\label{fig:values:boundary}}[0.23\textwidth]{
        {\includegraphics[width=0.23\textwidth]{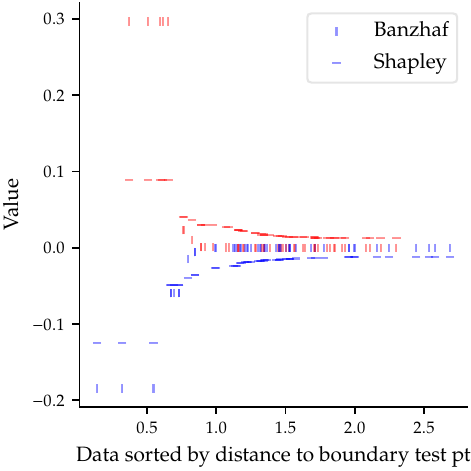}}
    }
    \quad
    \subcaptionbox{The values of the data points with respect to a test point that is far from the boundary.\label{fig:values:far}}[0.23\textwidth]{
        {\includegraphics[width=0.23\textwidth]{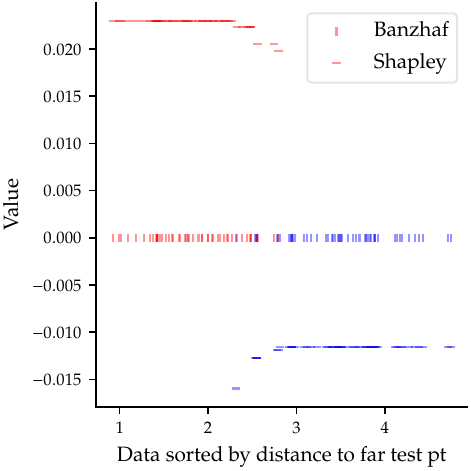}}
    }
    \quad
    \subcaptionbox{\revise{Injected random noise (upper-left corner) to the 2D dataset, with test points in square.}\label{fig:values:inject}}[0.23\textwidth]{
        {\includegraphics[width=0.23\textwidth]{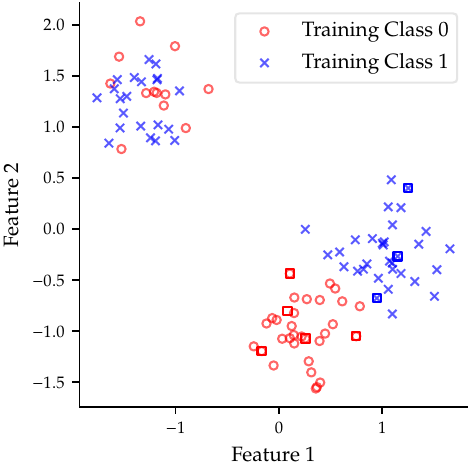}}
    }

    \caption{The difference between the Shapley and Banzhaf values. The model performance is measured by the accuracy of a \knn classifier with $k=5$.}
    \label{fig:values}
\end{figure*}

We illustrate the major differences between different values in \cref{fig:values}.
The values of all data points in a 2D dataset (\cref{fig:values:dataset}) are computed with respect to a boundary test point (square) and a test point that is far from the boundary (triangle).
\revise{Only three training data points have non-zero LOO values.
After duplicating the dataset by creating a copy of each original data instance, all LOO values become zero.
In contrast, most non-zero game-theoretic values remain non-zero and informative after duplication.
This simple observation suggests that LOO values are less informative, especially in the presence of data redundancy~\citep{lee2022deduplicating,bousquet2002stability,hardt2016train}.}

\rrevise{Next, we compare the Shapley and Banzhaf values to illustrate why they can behave differently and be preferable in different applications.}
\revise{There are three key observations in \cref{fig:values}.}
First, the Banzhaf values are \emph{sparser} than the Shapley values, while the Shapley values remain far from zero even for data points that are very far from the test point (\cref{fig:values:boundary}).
This sparsity property can be very helpful in some applications, such as exemplar-based interpretability~\citep{rudin2022interpretable}.
Second, for a test point (triangle) that is surrounded by a majority of data points of its own class, the Banzhaf values of these neighboring points are close to zero (\cref{fig:values:far}).
Intuitively, this means that removing any of these neighboring points has little impact on the prediction of the test point.
Thus, the Banzhaf values can not only highlight more concentrated relative importance of the data points to a boundary point, but also indicate whether the test point has been well-supported, and there is no need for further data enhancement.
\revise{Third, the Banzhaf values are more robust to the noise in the dataset (\cref{fig:values:inject}), where additional random-label training data points are injected (upper-left corner).
The Banzhaf values of the noise are all zero, and the values of the original data points remain unchanged.
In contrast, the total Shapley value of the original data points drops by over 25\% due to the non-zero Shapley values of the noise data.
The fragility of the Shapley values originates from its over\-allocation of values to extreme subsets;
see definition in~\cref{eq:sval-def-weighted}.
Robustness is crucial in adversarial settings where malicious parties may inject useless or mislabeled data to dilute the value of legitimate contributors and siphon off monetary rewards in data marketplaces, 
or in environments with noisy data.}

Despite the theoretical appeal of these game-theoretic approaches to data valuation, their practical application faces a significant obstacle: computational complexity. 
Computing exact Shapley or Banzhaf values requires evaluating the value function over all possible subsets of the dataset, leading to exponential complexity. %
In general, it has been proved that both Shapley and Banzhaf values are \sharpp-hard to compute~\citep{deng1994complexity}.
This computational burden becomes prohibitive for real-world datasets, which often contain thousands or millions of samples. 
Consequently, most existing work has focused on approximation methods, such as sampling-based approaches~\citep{fatima2008linear,castro2009polynomial,bachrach2010approximating,maleki2013bounding,mitchell2022sampling,zhang2023efficient}, which trade accuracy for efficiency.

In this paper, we address the challenge by developing fast dynamic programming (DP) algorithms \rrevise{specifically tailored for computing Banzhaf values in $k$-nearest neighbor (\knn) \emph{classifiers}.}
The classic \knn classifiers remain one of the most central and widely-used models in machine learning, especially when combined with recent pre-trained embedding techniques.
Our approach exploits the locality property of \knn and computes exact Banzhaf values efficiently without resorting to approximation.
\rrevise{Our algorithmic contributions target \knn classification with \emph{hard-label} valuation (see definition in \cref{sec:prelim:knn-bval}).
We include soft-label and Shapley variants in experiments for comparison.
We offer the fastest algorithms for hard-label Banzhaf valuation, which are even more efficient than their hard-label Shapley counterparts.}
Our contribution bridges the gap between theoretical foundations of cooperative game theory and practical demands of real-world machine-learning applications.
In detail, this paper makes the following contributions:
\begin{itemize}
    \item We prove that it is \sharpp-hard to compute the Banzhaf values for the \knn classification model.
    
    \item We present the first dynamic programming algorithm for computing exact Banzhaf values in weighted \knn classifiers in pseudo-polynomial $\bigO(Wn^4)$ time, where $W$ is the maximum sum of top-$k$ weights. 
    Our algorithm enables exact computation of Banzhaf values without approximation.
    
    \item We then significantly improve the time complexity of the DP algorithm for computing exact Banzhaf values in weighted \knn classifiers to $\bigO(Wkn^2)$, through novel algorithmic techniques including efficient partial sum computation and recurrence relations for binomial coefficients.
    
    \item We present a specialized efficient algorithm for unweighted \knn %
    achieving $\bigO(nk^2)$ time complexity. 
    Thus, the running time of our algorithm is essentially \emph{linear} in the number of training points, 
    since $k$ is typically a small constant in practice.
    To the best of our knowledge, this is the first near-linear time algorithm for computing the (hard-label) \knn-based game-theoretic values.
    
    \item We also provide efficient coalition-based and permutation-based sampling algorithms for computing approximate Banzhaf values, by exploiting the locality property of \knn classifiers.
        
    \item Our experimental evaluation demonstrates the practical efficiency of our algorithms on real-world datasets, and their effectiveness in various applications such as data acquisition and cleaning.
    \rrevise{In particular, the unweighted exact DP algorithm scales to much larger training sets, while the weighted exact DP algorithm is more efficient than Shapley-value baselines under the same hard-label setting.}
\end{itemize}

\rrevise{We note that our framework does not directly extend to \knn regression. The standard regression utility introduces a normalization term that is not decomposable under our DP recurrence, and the continuous output eliminates the pivotal-subset counting structure used in hard-label classification. Therefore, regression is beyond the scope of this paper (see \cref{sec:dp:practical:regression} for details).}

The rest of the paper is organized as follows.
We begin by introducing the necessary preliminaries in \cref{sec:preliminaries}.
Then, we present our dynamic programming and sampling algorithms in \cref{sec:alg} and \cref{sec:mc}, respectively.
The related work is discussed in \cref{sec:related} and our experimental evaluation is presented in \cref{sec:experiments}.
We conclude in \cref{sec:conclusion}.

\section{Preliminaries}
\label{sec:preliminaries}
We first define the Banzhaf and Shapley values in the general framework of cooperative game theory.
Then, we introduce the Banzhaf values for the \knn classification model and we prove its hardness.

\subsection{Banzhaf and Shapley values}
\label{sec:prelim:bval-sval}

The Banzhaf values are a fundamental concept in cooperative game theory.
They provide a principled approach to distribute in a fair manner the total value generated by a coalition over its members. 
Originally introduced by \citet{banzhaf1965weighted} in the context of voting power analysis, the idea has found widespread applications in various domains, including economics, political science, and more recently, machine learning for data-valuation problems.

To formally define the Banzhaf values, we begin with the standard framework of cooperative games. 
A cooperative game is characterized by a pair $(N, v)$, where $N = \{1, 2, \ldots, n\}$ represents the set of players and $v: 2^N \to \mathbb{R}$ is a %
function that assigns a real value to each subset (coalition) $S \subseteq N$ of players. 
The value %
$v(S)$ represents the total value of the coalition $S$. 
By convention, we assume $v(\emptyset) = 0$, meaning that the empty coalition generates no~value.

For any player $i \in N$ and coalition $S \subseteq N \setminus \{i\}$, the \emph{marginal contribution} of player $i$ to coalition $S$ is defined as $v(S \cup \{i\}) - v(S)$. 
This quantity captures the additional value generated when player $i$ joins an existing coalition $S$. 
The Banzhaf value of player $i$, denoted $\bval_i(v)$, is then defined as the expected marginal contribution of player $i$ over all possible coalitions, that is,
\begin{equation}
    \bval_i(v) = \frac{1}{2^{n-1}} \sum_{S \subseteq N \setminus \{i\}} [v(S \cup \{i\}) - v(S)]. \label{eq:bval-def}
\end{equation}

For comparison, we also define the Shapley value $\sval_i(v)$ for player~$i$, 
which is defined as 
\begin{equation}
    \sval_i(v) = \frac{1}{n!} \sum_{\pi \in \Pi(N)} \left[ v(\pi_i \cup \{i\}) - v(\pi_i) \right], \label{eq:sval-def}
\end{equation}
where $\Pi(N)$ denotes the set of all permutations of $N$, and $\pi_i$ denotes the set of players that precede player $i$ in $\pi$~\citep{shapley1953value}.
Notably, the Shapley value can be rewritten as the weighted average over all possible coalitions, as follows:
\begin{equation}
    \sval_i(v) = \frac{1}{n} \sum_{S \subseteq N \setminus \{i\}} \binom{n-1}{|S|}^{-1} \left[ v(S \cup \{i\}) - v(S) \right]. \label{eq:sval-def-weighted}
\end{equation}

Unlike the Banzhaf value that treats all coalitions equally, the Shapley value puts more emphasis on the marginal contribution of a player to 
extremely small and large coalitions, while under-emphasizing those of medium sizes.

\subsection{Banzhaf value for \knn classifier}
\label{sec:prelim:knn-bval}

In the context of machine learning, we adapt the Banzhaf value framework to quantify the contribution of individual training data points to model performance. 
Consider a training dataset $D = \{z_1,\ldots, z_n\}$ where each data point $z_i = (x_i, y_i)$ 
represents a feature-label pair with $y_i \in \mathcal{Y}$, 
that is, $\mathcal{Y}$ is the set of labels.
For a given test data point $\ztest = (\xtest, \ytest)$, we define the players in our cooperative game as the training data points.
We specifically focus on the \knn classifier setting.

The valuation function $v$ for the \knn classifier is defined as follows. 
For any subset $S \subseteq D$ of training data points and a fixed test point $\ztest$, 
the valuation function $v(S \mid \ztest)$ represents the accuracy of the \knn classifier 
trained on the subset $S$ when predicting the label of the test point, which is
\begin{equation}
    v(S \mid \ztest) = \begin{cases}
    1 & \text{if } \ytest = \arg\max_{y \in \mathcal{Y}} \sum_{i=1}^{\min\{|S|, k\}} w_i \, \mathbb{1}\{y_i(S) = y\} \\
    0 & \text{otherwise,}
    \end{cases}
    \label{eq:knn-value-function}
\end{equation}
where $z_i(S)$ is the $i$-th closest point of $S$ to $\xtest$, and $w_i = w(z_i(S) \mid \ztest)$ is the weight of $z_i(S)$.
Note that the valuation function is zero when $|S| = 0$. 
We also define it to be zero when the model predicts more than one label,
which is possible because of ties.
For the \knn classifier, the weight $w_i$ is determined by a non-increasing function 
of the distance between $z_i(S)$ and $\xtest$~\citep{gyorfi2002distribution}, 
or simply set to $w_i=1$ for the unweighted \knn.
Without loss of generality, we assume that the weights are integers.

In the case of binary classification, i.e., $|\mathcal{Y}| = 2$, we can further simplify the valuation function as follows.
We focus on this case in the paper for simplicity, and discuss extension to the general case in \cref{sec:dp:practical:multiclass}.
Then, the valuation function is given by
\begin{equation}
    v(S \mid \ztest) = \begin{cases}
    1 & \text{if } \sum_{i=1}^{\min\{|S|, k\}} w_i\, (-1)^{\mathbb{1}\{y_i(S) \neq \ytest\}} > 0 \\
    0 & \text{otherwise},
    \end{cases}
    \label{eq:knn-value-function-binary}
\end{equation}
where  $w_i = w(z_i \mid \ztest)$ and where
the factor $(-1)^{\mathbb{1}\{y_i(S) \neq \ytest\}}$ becomes~$-1$
if $y_i \neq \ytest$, or $+1$ otherwise.
To keep the notation simple, we re-define $w(z_i \mid \ztest)$ as $w(z_i(S) \mid \ztest) (-1)^{\mathbb{1}\{y_i(S) \neq \ytest\}}$ for binary classification in subsequent sections.

With the valuation function defined, the Banzhaf value of a training data point $z_i$ is given by
\begin{equation}
    \bval_i(v \mid \ztest) = \frac{1}{2^{n-1}} \sum_{S \subseteq N \setminus \{i\}} [v(S \cup \{i\} \mid \ztest) - v(S \mid \ztest)].
    \label{eq:bval-knn}
\end{equation}
We often omit the test point $\ztest$ in the notation for simplicity when the context is clear.

When provided with a test dataset $\Dtest$ comprising $\ntest$ test data points, the final Banzhaf value of a training data point $z_i\in D$ is the average of the Banzhaf values over all test points:
\begin{equation}
    \bval_i(v \mid \Dtest) = \frac{1}{\ntest} \sum_{\ztest \in \Dtest} \bval_i(v \mid \ztest).
    \label{eq:bval-knn-avg}
\end{equation}

\begin{remark}
The valuation function $v(S \mid \ztest)$ defined in \cref{eq:knn-value-function} is also known as the \emph{hard-label} valuation function.
In contrast, there exists a \emph{soft-label} counterpart~\citep{wang2023note} that is defined as follows.
\begin{equation}
	v(S \mid \ztest) = 
	\begin{cases}
		\frac{\sum_{i=1}^{\min\{|S|, k\}} w(z_i(S) \mid \ztest) \, \indicator(y_i(S) = \ytest)}{\sum_{i=1}^{\min\{|S|, k\}} w(z_i(S) \mid \ztest)} & \text{if } |S| > 0 \\
		1/|\mathcal{Y}| & \text{if } |S| = 0.
		\label{eq:knn-value-function-soft}
	\end{cases}
\end{equation}
The soft-label valuation function captures a more fine-grained notion of contribution, while the use of the hard-label valuation function identifies the most decisive players.
We will demonstrate in the experiments that both versions are useful in different scenarios.
\end{remark}

\subsection{Hardness of computing Banzhaf values for \knn classifier}
\label{sec:prelim:hardness}

We now prove that computing the Banzhaf values for the \knn classifier, 
as defined in the previous section, is a \sharpp-hard problem.
We prove this hardness result via a reduction from the \textsc{Weighted-Majority-Game} ({\small WMG}) problem~\citep{prasad1990np}, a classical \sharpp-hard problem in cooperative game theory.
Our hardness result suggests that, under standard complexity assumptions, there is unlikely to exist a (strongly) polynomial-time algorithm for computing Banzhaf values for \knn classifier.

\begin{theoremE}\label{thm:hardness}
	Given a training dataset $D = \{z_1,\ldots, z_n\}$ and a test data point $\ztest$, 
	computing the Banzhaf values $\bval_i(v \mid \ztest)$ for \knn classifier, as in \cref{eq:bval-knn}, 
	for an arbitrary $z_i \in D$ is $\sharpp$-hard.
\end{theoremE}
\begin{proofE}
	We use a reduction from the \textsc{Weighted-Majority-Game} ({\small WMG}) problem, 
	which is defined as follows. 
	The input consists of a set of players $N' = \{1, \dots, n'\}$ 
	with non-negative integer weights $w_1, \dots, w_{n'}$, an integer quota $q > 0$, and a designated player $p \in N'$.
	The goal is to compute the number of \emph{swing coalitions} for player $p$, denoted by $\eta_p$. 
	A subset $S' \subseteq N' \setminus \{p\}$ is a swing coalition if
	\[
	\sum_{j \in S'} w_j < q
	\quad\text{and}\quad
	\sum_{j \in S'} w_j + w_p \ge q .
	\]
	It has been shown that computing $\eta_p$ is \sharpp-complete~\citep{prasad1990np}.
	
	Given an instance of the {\small WMG} problem, we construct an instance of the \knn-Banzhaf problem 
	in polynomial time as follows.
	First, we take $\mathcal{Y}=\{+1,-1\}$ and create a single test point $\ztest$ with label $\ytest=+1$. 
	Next we construct a training dataset $D$ of size $|D| = n = n' + 1$.
	For each {\small WMG} player $j \in N'$, we create a training data point $z_j \in D$ 
	with label $y_j = \ytest = +1$ and an integer weight equal to the {\small WMG} weight, $w(z_j) = w_j$.
	We additionally create one dummy point $z_d \in D$ with %
	label $y_d =-1$ and weight $w(z_d) = -(q-1)$. 
	We set the parameter $k = n = n' + 1$ and fix point $z_p$ whose Banzhaf value 
	we compute corresponding to {\small WMG} player $p$. 
	The size of the \knn instance is polynomial in the size of the {\small WMG} instance, and the reduction takes polynomial time.

	For the correctness of the reduction, we can show that 
	computing the value $\bval_p$ for a data point $z_p$ in the \knn game is equivalent to 
	computing $\eta_p$ for the corresponding player $p$ in the {\small WMG} game. 
	Let $S' \subseteq N' \setminus \{p\}$ be an arbitrary coalition of {\small WMG} players. 
	In our \knn construction, this is associated with the set of training points $S = \{z_j \mid j \in S'\} \cup \{z_d\}$. 
	We analyze the marginal contribution of $z_p$ to the set $S$.
	First, consider the \knn prediction for the coalition $S$ without $z_p$. 
	The total weight is 
	$\left(\sum_{j \in S'} w_j\right) + w(z_d) = \left(\sum_{j \in S'} w_j\right) - (q-1)$. 
	According to the valuation function, we have 
	$$ v(S) = 1 ~\Leftrightarrow~ \sum_{j \in S'} w_j - (q-1) > 0 ~\Leftrightarrow~ \sum_{j \in S'} w_j \ge q.$$
	For a swing to occur, $S$ must be a ``losing'' coalition, so we need $v(S)=0$, which holds if and only if $\sum_{j \in S'} w_j < q$.
	Next, consider the \knn prediction for the coalition $S \cup \{z_p\}$. 
	The total weight is $\left(\sum_{j \in S'} w_j + w_p\right) - (q-1)$ and
	\begin{align*}
		v(S \cup \{z_p\}) = 1 
			& ~\Leftrightarrow~ \sum_{j \in S'} w_j + w_p - (q-1) > 0 \\
			& ~\Leftrightarrow~ \sum_{j \in S'} w_j + w_p \ge q.
	\end{align*}
	
	Therefore, the point $z_p$ is pivotal for the coalition $S$, i.e., $v(S \cup \{z_p\}) - v(S) = 1$, if and only if both of the {\small WMG} swing conditions are met. 
	This establishes a one-to-one correspondence between the swing coalitions for player $p$ in the {\small WMG} and the pivotal coalitions for point $z_p$ in the constructed \knn game. 
	The number of such pivotal coalitions denoted by $\mu_{z_p}$ is thus identical to $\eta_p$.
	
	The Banzhaf value is defined as $\bval_{z_p} = \frac{1}{2^{n-1}} \mu_{z_p}$. 
	Our reduction shows that this is equal to $\frac{1}{2^{n-1}} \eta_p$. 
	If an oracle could compute $\bval_{z_p}$ in polynomial time, we could compute the \sharpp-complete value $\eta_p$ via $\eta_p = \bval_{z_p} \cdot 2^{n-1}$. 
	This implies that computing $\bval_{z_p}$ is \sharpp-hard.
\end{proofE}

\section{Efficient computation of Banzhaf values for \knn classifier}
\label{sec:alg}

We first describe how to compute the Banzhaf values of all training points in $\bigO(n^4 W)$ time using a dynamic programming (DP) approach, where $W$ is the maximum sum of the top-$k$ weights.
As mentioned before, we assume a single test data point $\ztest$.
Then, we improve the time complexity to $\bigO(Wkn^2)$ by introducing many novel ideas.
Finally, we develop a more efficient algorithm for unweighted \knn with $\bigO(n k^2)$ time complexity.
Note that in practice $k$ is typically small, such as a value smaller than 10, 
so the time complexity is essentially linear in $n$.
\revise{We defer the discussion about practical implementation, and extension to multi-class and regression settings to \cref{sec:dp:missing}\citefullpaper.}

\subsection{A standard DP algorithm}
\label{sec:alg:standard}

Following the definition of Banzhaf values in \Cref{eq:bval-def},
to compute $\bval_i(v)$ for any $z_i \in D$ we need to count the number of subsets $S \subseteq D \setminus \{z_i\}$ for which $z_i$ is \emph{pivotal}, that is, the addition of $z_i$ changes the outcome of the game.
Interestingly, we can avoid enumerating all (exponentially many) subsets $S$
by performing the counting via \emph{dynamic programming}. 

Before presenting the counting algorithm, it is useful to determine in advance what the possible outcomes of adding a data point $z_i$ to a pivotal subset are.
\begin{lemmaE}
    \label{lem:pivotal-outcomes}
    If $y_i = \ytest$, it is impossible for $z_i$ to turn $v(S)=1$ into $v(S \cup \{z_i\})=0$.
    Similarly, if $y_i \neq \ytest$, it is impossible for $z_i$ to turn $v(S)=0$ into $v(S \cup \{z_i\})=1$.
\end{lemmaE}
\begin{proofE}
    We only discuss the case when $y_i = \ytest$, and the proof for the other case is similar.
    A point with a larger weight is closer to $\ztest$.
    Thus, $z_i$ cannot push a point with a larger weight out of the top-$k$ neighbors to turn $v(S)=1$ into $v(S \cup \{z_i\})=0$.
\end{proofE}

We now describe the dynamic-programming approach for the counting task.
Given a test point $\ztest$, we rename the data points in the training dataset $D$ so that $z_1, \ldots, z_n$ are sorted by non-decreasing distance to $\ztest$.
Let $D_{\bar{i}} = D \setminus \{z_i\}$.

We define a function $f_i(w, s, m)$ to represent the number of subsets $S \subseteq D_{\bar{i}}$ such that
(1)~$|S| = s$;
(2)~the sum of top-$\min\{s, k\}$ weights equals $w$; and
(3)~the $\min\{s, k\}$-th closest point of $S$ to $\ztest$ is $z_m$.
Mathematically,
\begin{align}
\begin{split}
    \label{eq:dp:f}
    &f_i(w, s, m) = \\ &\sum_{S \subseteq D_{\bar{i}}} \mathbb{1} \left( |S| = s \text{ and } \sum_{i=1}^{\min\{s, k\}} w(z_i(S)) = w \text{ and } z_{\min\{s, k\}}(S) = z_m \right).
\end{split}
\end{align}
Recall that $z_i(S)$ is the $i$-th closest point to $\ztest$ among the data points in $S$. 
We will see shortly that the Banzhaf value $\bval_i(v)$ can be obtained by aggregating the values of $f_i$.
To efficiently compute $f_i$, we derive the following recurrence.
\begin{lemmaE}
    \label{lem:dp:standard}
    If $s \le k$, then
\[
    f_i(w, s, m) = \sum_{m' < m: m' \ne i} f_i(w - w(z_m), s-1, m'),
\]
    else, if $m > i$, then
\[
    f_i(w, s, m) = f_i(w, k, m) \cdot \binom{n-m}{s-k}.
\]
\end{lemmaE}
\begin{proofE}
    We choose to construct $f_i$ by gradually increasing the subset size $s$.
    We first discuss the case when $s \le k$.
    For a subset $S$ to \emph{end} with $z_m$, that is, $z_{\min\{s, k\}}(S) = z_m$,
    the subset $S \setminus \{z_m\}$ must end with a point $z_m'$ that is ranked higher than $z_{m}$, i.e., $m' < m$.
    This establishes the first recurrence relation.
    When $s > k$, note that for any point $z_m$, inserting any point $z_m'$ with $m' > m$ will only increase the subset size by one but will not violate the other constraints.
    Therefore, to increase the subset size from $k$ to $s$, one just needs to count how many possible combinations of $(s-k)$ different such points $z_m'$ exist, which turns out to be $\binom{n-m}{s-k}$.
\end{proofE}
Note that when $s > k$, there is no need to compute $f_i(w, s, m)$ for $m < i$, as inserting $z_i$ into those subsets has no effect.
We choose to construct $f_i$ by gradually increasing the subset size $s$.
We start with a special base case, $f_i(0, 0, 0) = 1$, which will be useful for $y_i = \ytest$ as $v(\emptyset) = 0$.
The base case for $s = 1$ is $f_i(w(z_m), 1, m) = 1$ for all $m \ne i$, and the rest values are initialized to zero.
Then, we fill the DP table of $f_i$ by enumerating the weight values from $-W$ to $W$, the subset size $s$ from 2 to $n-1$, and $m$ from 1 to $n$ with $m \ne i$.

It is not hard to see that the time complexity is $\bigO(W n^3)$ for each $f_i$, as each entry may take $\bigO(n)$ time.
Note that for $s > k$, computing the binomial coefficient also requires $\bigO(n)$ time.
Hence, computing all values of $f_i$ results in $\bigO(W n^4)$ time in total.

Once the values of $f_i$ have been computed, we can obtain the Banzhaf value of $z_i$ with $y_i = \ytest$ by
\begin{equation}
    \bval_i(v) = \frac{1}{2^{n-1}} \left( C_{<k} + C_{\ge k} \right),
    \label{eq:dp:bval-positive}
\end{equation}
where 
\[
    C_{<k} = \sum_{\substack{w \in (-w(z_i),0] \\ s \in [0,k-1]\\ m \in [0,n]:m \ne i}} f_i(w, s, m),
\]
and
\[
    C_{\ge k} = \sum_{\substack{w \in (-w(z_i)+w(z_m),0]\\ s \in [k,n-1]\\ m > i}} f_i(w, s, m).
\]
Note that we use the notation $[a,b]$ to denote a closed interval in the set of integers here.
Both terms count the number of valid subsets $S$ where $z_i$ is pivotal.
By \cref{lem:pivotal-outcomes}, a positive $z_i$ can only turn the outcome value from $0$ to~$1$.
Here $C_{<k}$ counts the number of subsets $S$ whose size is less than $k$, and $C_{\ge k}$ counts the number of those whose size is at least~$k$.
The difference is that for $C_{\ge k}$, inserting $z_i$ will push the current $k$-th closest point $z_m$ out of the top-$k$ neighbors.
Therefore, the range of weight where $z_i$ can be pivotal has to be adjusted according to~$w(z_m)$.

Similarly, if $y_i \neq \ytest$, then
\begin{equation}
    \bval_i(v) = -\frac{1}{2^{n-1}} \left( C_{<k} + C_{\ge k} \right), \label{eq:dp:bval-negative}
\end{equation}
where
\[
    C_{<k} = \sum_{\substack{w \in (0, -w(z_i)]\\ s \in [0,k-1]\\ m \in [1,n]:m \ne i}} f_i(w, s, m),
\]
and
\[
    C_{\ge k} = \sum_{\substack{w \in (0, -w(z_i)+w(z_m)]\\ s \in [k,n-1]\\ m > i}} f_i(w, s, m).
\]
One difference from the positive case is that the empty set $S$ represented by $f_i(0, 0, 0)$ is excluded in $C_{<k}$.

\begin{algorithm}[t] %
	\DontPrintSemicolon
	\KwIn{Dataset $D$, test point $\ztest$, integer $k$}
	\SetKwComment{tcp}{$\triangleright$\ }{}%
    \SetKwFunction{FuncBanzhafDP}{BanzhafDP}
	\SetCommentSty{small}

    Let $z_1, \ldots, z_n$ be the data points sorted by non-decreasing distance to $\ztest$\;
    Let $W$ be the maximum sum of top-$k$ weights\;
    
    \For{$i = 1$ \KwTo $n$}{
        $\bval_i(v) \leftarrow$ $\FuncBanzhafDP(i, \{z_1, \ldots, z_n\}, k, W)$\;
    }
    \Return{$\bval_1(v), \ldots, \bval_n(v)$}\;    
	\caption{DP framework for \knn-Banzhaf values}
	\label{alg:dp:framework}
\end{algorithm}

\begin{algorithm}[t] %
	\DontPrintSemicolon
	\SetKwComment{tcp}{$\triangleright$\ }{}%
    \SetKwFunction{FuncBanzhafDP}{BanzhafDP}
	\SetCommentSty{small}

	\Function{$\FuncBanzhafDP(i, \{z_1, \ldots, z_n\}, k, W)$}{
        \tcp{Initialize the DP table} 
        $f_i(w, s, m) \leftarrow 0$ for all $w \in [-W, W]$, $s \in [0, n-1]$, $m \in [1, n]$\;

        $f_i(0, 0, 0) \leftarrow 1$\;
        
        \For{$m = 1$ \KwTo $n$ where $m \neq i$}{
            $f_i(w(z_m), 1, m) \leftarrow 1$\;
        }
        
        \tcp{fill the DP table} 
        \For{$s = 2$ \KwTo $n-1$}{
            \For{$w = -W$ \KwTo $W$}{
                \For{$m = 1$ \KwTo $n$ where $m \neq i$}{
                    \If{$s \leq k$}{
                        $f_i(w, s, m) \leftarrow \sum_{m' < m: m' \ne i} f_i(w - w(z_m), s-1, m')$\;
                    }
                    \Else{
                        \If{$m > i$}{
                            $f_i(w, s, m) \leftarrow f_i(w, k, m) \cdot \binom{n-m}{s-k}$\;
                        }
                    }
                }
            }
        }
        \If{$y_i = \ytest$}{
            \Return{$\frac{1}{2^{n-1}} \left( C_{<k} + C_{\geq k} \right)$ according to \cref{eq:dp:bval-positive}}\;
        }
        \Else{
            \Return{$-\frac{1}{2^{n-1}} \left( C_{<k} + C_{\geq k} \right)$ according to \cref{eq:dp:bval-negative}}\;
        }
    }
    
	\caption{Standard dynamic programming algorithm}
	\label{alg:dp}
\end{algorithm}

The algorithm is shown in \cref{alg:dp} with an example illustrated in \cref{ex:dp:standard}.
\Cref{alg:dp:framework} invokes \cref{alg:dp} for each data point $z_i\in D$.
We summarize the time complexity of our first non-optimized algorithm as follows.
\begin{theorem}[Standard DP]
    \label{thm:dp:standard}
    The dynamic programming algorithm in \cref{alg:dp:framework,alg:dp} for computing the \knn-Banzhaf values of all data points runs in $\bigO(Wn^4)$ time, where $W$ is the maximum sum of the top-$k$ weights and $n$ is the number of data points in $D$.
\end{theorem}

\subsection{A more efficient DP algorithm}
\label{sec:alg:efficient}

While the dynamic-programming algorithm we presented in the previous section has 
pseudo-polynomial complexity, its running time is prohibitively expensive in practice
for any other than small datasets.
To improve the running time, we introduce several novel techniques, 
including pre-computing partial sums and exploiting the recurrence relation of the binomial coefficients.
The result is an improved version of the DP algorithm with running time~$\bigO(Wkn^2)$. 

In our analysis, we distinguish two cases, $s \le k$ and~$s > k$.

\subsubsection{Run time improvement for $s \le k$}
In the standard DP, the time complexity of computing $C_{<k}$ is $\bigO(Wkn^2)$ for each $f_i$.
We show how to improve it to $\bigO(Wkn)$ by pre-computing partial sums.

For each term $f_i(w, s, m)$, we need to compute the sum of $f_i(w-w(z_m), s-1, m')$, for $m' < m$.
This summation can be sped up if we pre-compute the partial sums for each $w$ and $s$.
In particular, we define a partial sum
\[
    M_i(w,s,m) = \sum_{m' < m: m' \ne i} f_i(w, s, m').
\]
Note that for fixed values of $w$ and $s$, the partial sums $M_i(w,s,m)$ can be computed in $\bigO(n)$ time  for all values of $m$, by the recurrence 
\[
    M_i(w,s,m+1) = M_i(w,s,m) + f_i(w, s, m).
\]
Then, for $s \le k$, we can rewrite $f_i(w, s, m)$ in \cref{lem:dp:standard} as
\begin{equation}
    f_i(w, s, m) = M_i(w-w(z_m), s-1, m).
    \label{eq:dp:f-improved:leqk}
\end{equation}
By the above modification, we can compute each $f_i(w, s, m)$ entry in $\bigO(1)$ time, at the expense of pre-computing $M_i(w, s, m)$ in $\bigO(Wkn)$ time for all $w$, $m$ and $s \le k$.
Thus, the time complexity of $C_{<k}$ reduces to $\bigO(Wkn)$ for each $f_i$, and $\bigO(Wkn^2)$ for all $f_i$.

\subsubsection{Run time improvement for $s > k$}

To accelerate the computation of $C_{\ge k}$, our first key observation is that $C_{\ge k}$ does not require access to individual $f_i(w, s, m)$ terms but their sum.
Therefore, %
it is sufficient to compute the sum of $f_i$'s for $s \ge k$.
Formally, we define
\begin{align*}
    F_i(w,m) &= \sum_{s \in [k,n-1]} f_i(w, s, m) 
    = \sum_{s \in [k,n-1]} f_i(w, k, m) \cdot \binom{n-m}{s-k} \\
    &= f_i(w, k, m) \cdot g(m),
\end{align*}
where we set $g(m) = \sum_{j \in [0,n-k-1]} \binom{n-m}{j}$.
We will see shortly that to compute $C_{\ge k}$, it is sufficient to compute $F_i(w,m)$ for all $w$ and $m$.
Thus, if we can devise a fast way to compute $g(m)$ for all $m$, we can also improve the time complexity of $C_{\ge k}$.
Note that $g(m)$ is independent of $i$, so we only need to compute it once for all $i$.
Once we have pre-computed $g(m)$, each $F_i(w,m)$ entry can be computed in $\bigO(1)$ time.

We exploit the recurrence relation of the binomial coefficients with different $j$ to compute $g(m)$ efficiently, that is,
\[
    \binom{n-m}{j+1} = \binom{n-m}{j} \frac{n-m-j}{j+1}.
\]
Therefore, fixing $m$, $g(m)$ can be computed in $\bigO(n)$ time, as each summand costs $\bigO(1)$ time by using the above recurrence relation.

Furthermore, we use the following recurrence for $g(m)$ across different values of $m$ that further reduces the time complexity.
\begin{lemmaE}
For the function $g(m) = \sum_{j \in [0,n-k-1]} \binom{n-m}{j}$ it holds that
    \label{lem:dp:g-recurrence}
    \[
        g(m) = 2 \cdot g(m+1) - \binom{n-m-1}{n-k-1}.
    \]
\end{lemmaE}
\begin{proofE}
    Starting from the definition \\$g(m) = \sum_{j=0}^{n-k-1} \binom{n-m}{j}$, we apply Pascal's identity $\binom{n-m}{j} = \binom{n-m-1}{j} + \binom{n-m-1}{j-1}$ to each binomial coefficient:
    \begin{align*}
        g(m) &= \sum_{j=0}^{n-k-1} \binom{n-m}{j} \\
        &= \sum_{j=0}^{n-k-1} \left[\binom{n-m-1}{j} + \binom{n-m-1}{j-1}\right] \\
        &= \sum_{j=0}^{n-k-1} \binom{n-m-1}{j} + \sum_{j=0}^{n-k-1} \binom{n-m-1}{j-1}.
    \end{align*}

    The first sum is exactly $g(m+1) = \sum_{j=0}^{n-k-1} \binom{n-m-1}{j}$.

    For the second sum, we substitute $j' = j-1$ (so $j = j'+1$):
    \begin{align*}
        \sum_{j=0}^{n-k-1} \binom{n-m-1}{j-1} &= \sum_{j'=-1}^{n-k-2} \binom{n-m-1}{j'} \\
        &= \sum_{j'=0}^{n-k-2} \binom{n-m-1}{j'},
    \end{align*}
    where we used the fact that $\binom{n-m-1}{-1} = 0$.

    Now, observe that:
    \begin{align*}
        g(m+1) &= \sum_{j=0}^{n-k-1} \binom{n-m-1}{j} \\
        &= \sum_{j=0}^{n-k-2} \binom{n-m-1}{j} + \binom{n-m-1}{n-k-1}.
    \end{align*}

    Therefore:
    \[
        \sum_{j=0}^{n-k-2} \binom{n-m-1}{j} = g(m+1) - \binom{n-m-1}{n-k-1}.
    \]

    Substituting back into our expression for $g(m)$:
    \begin{align*}
        g(m) &= g(m+1) + \left[g(m+1) - \binom{n-m-1}{n-k-1}\right] \\
        &= 2 \cdot g(m+1) - \binom{n-m-1}{n-k-1}.
    \end{align*}
\end{proofE}
\cref{lem:dp:g-recurrence} allows us to compute $g(m)$ for all $m$ in $\bigO(n)$ time in total, 
rather than $\bigO(n^2)$ if computed independently.
More specifically, we first compute $g(m)$ for the largest value of $m$ in $\bigO(n)$ time, and then use the recurrence to compute the next $g(m-1)$ values in $\bigO(1)$ time.
This is possible because as $m$ increases, the coefficient 
in \cref{lem:dp:g-recurrence} can be updated in $\bigO(1)$ time using the following identity:
\[
    \binom{n-(m+1)-1}{n-k-1} = \binom{n-m-1}{n-k-1} \cdot \frac{k-m}{n-m-1}.
\]
Actually, there are at most $k$ such coefficients, as $\binom{n-m-1}{n-k-1}=0$ for~$m > k$.
With the above adaptations, we now rewrite $C_{\ge k}$ when $y_i = \ytest$ as
\begin{equation}
    C_{\ge k} = \!\!\!\!\sum_{\substack{w \in (-w(z_i)+w(z_m),0]\\ m > i}}\!\!\!\! F_i(w, m).
    \label{eq:dp:efficient:c-geqk}
\end{equation}
The case for $y_i \neq \ytest$ is similar.
For each value of $i$, the time complexity of $C_{\ge k}$ is $\bigO(Wn)$ given pre-computed values of $g(m)$, 
which yields $\bigO(Wn^2)$ in total, for all $i$.
Hence, $C_{\ge k}$ is no longer the bottleneck of the DP algorithm.
The total time complexity for both~$C_{<k}$ and $C_{\ge k}$ reduces to $\bigO(Wkn^2)$, for all $i$.

\begin{algorithm}[t] %
	\DontPrintSemicolon
	\SetKwComment{tcp}{$\triangleright$\ }{}%
    \SetKwFunction{FuncBanzhafFastDP}{BanzhafFastDP}
	\SetCommentSty{small}

    \Function{$\FuncBanzhafFastDP(i, \{z_1, \ldots, z_n\}, k, W)$}{
        \tcp{Pre-compute $g(m)$ using recurrence relation in \cref{lem:dp:g-recurrence}}
        $g(n) \leftarrow \sum_{j=0}^{n-k-1} \binom{0}{j} = 1$\;
        \For{$m = n-1$ \KwTo $1$}{
            $g(m) \leftarrow 2 \cdot g(m+1) - \binom{n-m-1}{n-k-1}$\;
        }
        
        \tcp{Initialize the DP table} 
        $f_i(0, 0, 0) \leftarrow 1$\;
        
        \For{$m = 1$ \KwTo $n$ where $m \neq i$}{
            $f_i(w(z_m), 1, m) \leftarrow 1$\;
        }
        
        \tcp{Compute for case $s \leq k$ with partial sums optimization}
        \For{$s = 2$ \KwTo $k$}{
            \For{$w = -W$ \KwTo $W$}{
                $M_i(w, s-1, 1) \leftarrow 0$\;
                \For{$m = 2$ \KwTo $n$ where $m \neq i$}{
                    $M_i(w, s-1, m) \leftarrow M_i(w, s-1, m-1) + f_i(w, s-1, m-1)$\;
                }
            }
            
            \For{$w = -W$ \KwTo $W$}{
                \For{$m = 1$ \KwTo $n$ where $m \neq i$}{
                    $f_i(w, s, m) \leftarrow M_i(w - w(z_m), s-1, m)$\;
                }
            }
        }
        
        \tcp{Compute $F_i(w,m)$ for case $s > k$}
        \For{$w = -W$ \KwTo $W$}{
            \For{$m = i+1$ \KwTo $n$}{
                $F_i(w, m) \leftarrow f_i(w, k, m) \cdot g(m)$\;
            }
        }

        \If{$y_i = \ytest$}{
            \Return{$\frac{1}{2^{n-1}} \left( C_{<k} + C_{\geq k} \right)$ according to \cref{eq:dp:bval-positive,eq:dp:efficient:c-geqk}}\;
        }
        \Else{
            \Return{$-\frac{1}{2^{n-1}} \left( C_{<k} + C_{\geq k} \right)$ according to \cref{eq:dp:bval-negative,eq:dp:efficient:c-geqk}}\;
        }
    }
    
	\caption{Efficient dynamic programming algorithm}
	\label{alg:dp:efficient}
\end{algorithm}

The improved DP algorithm is displayed in \cref{alg:dp:efficient}, and its time complexity is summarized in the following theorem.

\begin{theorem}[Efficient DP]
    \label{thm:dp:efficient}
    The improved dynamic programming algorithm in \cref{alg:dp:framework,alg:dp:efficient} for computing the Banzhaf values for the \knn classifier of all data points runs in $\bigO(Wkn^2)$ time, where $W$ is the maximum sum of the top-$k$ weights, $k$ is the number of nearest neighbors, and $n$ is the number of data points in $D$.
\end{theorem}

\subsection{A near-linear time DP algorithm for unweighted \knn classifier}
\label{sec:alg:efficient-unweighted}

The improved DP algorithm presented in \cref{sec:alg:efficient},
although it brings a large improvement compared to standard DP,
it still requires quadratic time in $n$ and is not scalable for very large datasets.
Thus, as a further improvement, we propose a specialized DP algorithm for unweighted \knn with a time complexity of $\bigO(nk^2)$.
Recall that $k$ is usually small in practice, so the running time is essentially linear.

\subsubsection{Re-designing the state space}

The design of the specialized DP algorithm deviates significantly from the algorithm presented 
in \cref{sec:alg:efficient}.
In a nutshell, it relies on two novel ideas.
First, we show how to reduce the size of the state space of the DP table, 
which was previously parameterized by three variables in each function~$f_i$.
Second, we show how to avoid similar computations of $f_i$ for different values of~$i$. 

We now introduce our solution to the above two challenges.

Let $z_1, \ldots, z_n$ be the training data points sorted by non-decreasing distance to $\ztest$, 
and let $D_i = \{z_1, \ldots, z_i\}$ and $D^i = \{z_i, \ldots, z_n\}$.

We define two functions
\[
    f_i(w,s) = \sum_{S \subseteq D_i} \mathbb{1} \left( |S| = s \text{ and } \sum_{i=1}^{\min\{|S|, k\}} w(z_i(S)) = w \right),
\]
and 
\[
    b_i(w,s) = \sum_{S \subseteq D^i} \mathbb{1} \left( |S| = s \text{ and } \sum_{i=1}^{\min\{|S|, k\}} w(z_i(S)) = w \right).
\]
We call $f_i$ the \emph{forward function} and $b_i$ the \emph{backward function}.
Compared with the standard DP, we choose not to keep track of the $\min\{|S|, k\}$-th closest point (previously denoted as $z_m$) in $S$.
To make up for the missing information, we maintain two additional \emph{signed} backward functions, which are defined as follows:
\[
    b_i^-(w,t) = \sum_{S \subseteq D^i} \mathbb{1} \left( y_t(S) \ne \ytest \text{ and } \sum_{i=1}^{t} w(z_i(S)) = w \right),
\]
and
\[
    b_i^+(w,t) = \sum_{S \subseteq D^i} \mathbb{1} \left( y_t(S) = \ytest \text{ and } \sum_{i=1}^{t} w(z_i(S)) = w \right).
\]

The rationale is that in the unweighted \knn case it is not necessary to know exactly which point gets pushed out of the top-$k$ neighbors, when inserting $z_i$ into a subset $S$ of size larger than or equal to $k$. 
Instead, since the data points are unweighted, it is only necessary to know the sign of the data point.
In the signed backward functions, the label of the $t$-th closest point of subset~$S$ determines the sign of the subset.
Note that we do not require the subset $S$ to be of size exactly $t$ in the signed backward functions, and we only require the sum of its top-$t$ weights to be $w$.

We now describe the recurrence relations for efficient construction of the above functions.
We start with the forward and backward functions.

\begin{lemma}
    \label{lem:dp:unweighted}
    If $s \le k$, then
\[
    f_i(w, s) = f_{i-1}(w-w(z_i), s-1) + f_{i-1}(w, s),
\]
and 
\[
    b_i(w, s) = b_{i+1}(w-w(z_i), s-1) + b_{i+1}(w, s).
\]
\end{lemma}

We omit the proof as it is fairly straightforward.
Practically, we start with $f_i(0, 0) = 1$ for all $i \in [n]$, and $f_1(w(z_1), 1) = 1$; 
the rest of the entries are initialized to zero.
Then we process $f_2, \ldots, f_n$ in a forward manner.
The procedure for $b_i$ is similar except that it is done in a backward manner with an additional base case $b_{n+1}(0,0)=1$. %
By \cref{lem:dp:unweighted}, the time to construct $f_i$ and $b_i$, for all $i \in [n]$ and $s \le k$, is~$\bigO(Wnk)$.

We continue with the signed backward functions.

\begin{lemmaE}
    \label{lem:dp:unweighted-signed}
    For $t = 1$, we have
\[
b_i^-(w(z_i), 1) = \begin{cases}
	2^{n-i} + b_{i+1}^-(w(z_i), 1)
	& \text{ if } y_i \ne \ytest \\
	b_{i+1}^-(w(z_i), 1) & \text{ otherwise}.
\end{cases}
\]
    For $t \in [2,k]$, we have
\[
    b_i^-(w, t) = b_{i+1}^-(w-w(z_i), t-1) + b_{i+1}^-(w, t).
\]
    The update rules for $b_i^+$ are identical except for using the condition $y_i = \ytest$ when $t=1$.
\end{lemmaE}
\begin{proofE}
    It is important to note that since we work backward, every new $z_i$ is the closest point to $\ztest$ that is seen so far.
    When $t=1$, if we include $z_i$, then it automatically decides the sign of the subset.
    When its sign matches that of the function, it is feasible to include it.
    In that case, including any point behind it does not affect the sign and the weight of the top-$t$ points, so there are $2^{n-i}$ such subsets that satisfy the condition.

    Apart from the above special case, the rest of the update rules are simple.
    At every new $z_i$, there are precisely two options: to include it or not.
    Summing up the number of feasible subsets for each option leads to the update rules.
\end{proofE}
The update rules work backward from $i=n$ to 1 for each $t$ with an additional base case $b_{n+1}(0,0)=1$. 
Fixing a size $t$, for each $i$, we sum up the cases where $z_i$ is in the subset $S$ or not.
Since we work backward, the newly added $z_i$ is always the closest point to $\ztest$ in the current subset $S$.
This makes the case of $t=1$ special because if the current point $z_i$ is included, it automatically decides the sign of the subset.
Therefore, we need to check if its sign matches that of the function.
If the signs match and it is included, then further including any point behind it will not change the sign and top-$t$ weights, which means that there are $2^{n-i}$ feasible subsets.
By \cref{lem:dp:unweighted-signed}, the time to construct $b_i^-$ and $b_i^+$, for all $i \in [n]$ up to $t \le k$,
is $\bigO(Wnk)$.

\subsubsection{Computing the \knn-Banzhaf values via aggregation}

After the above preparations, we are ready to compute the \knn-Banzhaf values.
We point out that all points $z_i$ can share the above common states, as one can compute the \knn-Banzhaf values for all $z_i$ by aggregating the needed subsets $S \subseteq D_{\bar{i}}$ from $D_{i-1}$ and $D^{i+1}$ for each $z_i$.
That is, if $y_i = \ytest$, the Banzhaf value of $z_i$ can be computed as
\begin{equation}
    \bval_i(v) = \frac{1}{2^{n-1}} \left( C_- + C_{<k} \right),
    \label{eq:dp:unweighted:bval:positive}
\end{equation}
where 
\[
    C_- = \sum_{\substack{w \in (-W,W) \\ s \in [0,k-1]}} 
        \left( f_{i-1}(w, s) \cdot \sum_{\substack{w'+w \in [-1,0]\\ t = k-s}} b_{i+1}^-(w', t) \right)
\]
and
\[
    C_{<k} = \sum_{\substack{w \in (-W,W) \\ s \in [0,k-1]}} 
        \left( f_{i-1}(w, s) \cdot \sum_{\substack{w'+w = 0\\ s+t \in [0,k-1]}} b_{i+1}(w', t) \right).
\]

In each $C$ term, we combine a subset $S_1 \subseteq D_{i-1}$ and a subset $S_2 \subseteq D^{i+1}$ to form a valid subset $S = S_1 \cup S_2$ where $z_i$ is pivotal.
More specifically, we require the weight of $S$ to fall into a certain range so that $z_i$ can change the outcome.
Note that $|S_1|$ has to be smaller than $k$, or otherwise $z_i$ cannot be pivotal.
Besides, we need to distinguish two cases depending on the size of the subset $S$.
The first $C_-$ term handles subsets $S$ whose size is at least $k$, \emph{and} whose $k$-th closest point is negative.
This is because if the $k$-th closest point is positive, inserting the positive point $z_i$ will not change the value of the subset.
The second $C_{<k}$ term counts the number of valid subsets whose size is less than $k$.
Since the subset size is less than $k$, no point will be pushed out of the top-$k$ neighbors, so no signed functions are needed.

For each $\bval_i$, it costs $\bigO(Wk)$ time to compute $C_-$, but needs $\bigO(Wk^2)$ time to compute $C_{<k}$.
To speed up the computation, we pre-compute the partial sum of the backward function $b_i$ by letting
\[
    B_i(w, t) = \sum_{t' \in [0,t]} b_i(w, t'),
\]
which can be computed in $\bigO(Wnk)$ time for all $w$, $i$ and $t$.
Then, $C_{<k}$ can be rewritten as
\[
    C_{<k} = \sum_{\substack{w \in (-W,W)\\ s \in [0,k-1]}} \left( f_{i-1}(w, s) \cdot B_i(-w, k-1-s) \right),
\]
which can now be computed in $\bigO(Wk)$ time.

The case for $y_i \neq \ytest$ is similar by summing up $C_{+}$ and $C_{<k}$ as 
\begin{equation}
    \bval_i(v) = -\frac{1}{2^{n-1}} \left( C_{+} + C_{<k} \right),
    \label{eq:dp:unweighted:bval:negative}
\end{equation}
where 
\[
    C_{+} = \sum_{\substack{w \in (-W,W) \\ s \in [0,k-1]}} 
        \left( f_{i-1}(w, s) \cdot \sum_{\substack{w'+w \in [1,2]\\ t = k-s}} b_{i+1}^-(w', t) \right)
\]
and
\[
    C_{<k} = \sum_{\substack{w \in (-W,W) \\ s \in [0,k-1]}} 
        \left( f_{i-1}(w, s) \cdot \sum_{\substack{w'+w = 1\\ s+t \in [0,k-1]}} b_{i+1}(w', t) \right).
\]

Hence, the total time complexity of $\bval_i$ for all $i$ is $\bigO(Wnk)$.
Note that $W=\bigO(k)$ for unweighted \knn, so it is $\bigO(nk^2)$ in total.

\begin{algorithm}[t] %
	\DontPrintSemicolon
	\KwIn{Dataset $D$, test point $\ztest$, integer $k$}
	\SetKwComment{tcp}{$\triangleright$\ }{}%
	\SetCommentSty{small}

    Sort $z_1, \ldots, z_n$ by non-decreasing distance to $\ztest$\;
    $W \leftarrow k + 1$\;
    
    Compute forward functions $f_i(w,s)$ according to \cref{lem:dp:unweighted}\;
    Compute backward functions $b_i(w,s)$ according to \cref{lem:dp:unweighted}\;
    Compute signed backward functions $b_i^+(w,t)$ and $b_i^-(w,t)$ according to \cref{lem:dp:unweighted-signed}\;
    
    \For{$i = 1$ \KwTo $n$}{
        \If{$y_i = \ytest$}{
            $\bval_i(v) \leftarrow \frac{1}{2^{n-1}} (C_- + C_{<k})$ according to \cref{eq:dp:unweighted:bval:positive}\;
        }
        \Else{
            $\bval_i(v) \leftarrow -\frac{1}{2^{n-1}} (C_+ + C_{<k})$ according to \cref{eq:dp:unweighted:bval:negative}\;
        }
    }
    
    \Return{$\{\bval_1(v), \ldots, \bval_n(v)\}$}\;
    
	\caption{Specialized DP algorithm for unweighted \knn (brief version)}
	\label{alg:dp:unweighted}
\end{algorithm}

The specialized unweighted DP algorithm (brief version) is displayed in \cref{alg:dp:unweighted}.
The full version is deferred to \cref{sec:dp:unweighted:full}\citefullpaper due to space constraints.
The algorithm does not follow the framework in \cref{alg:dp:framework} because all data points share the same DP tables.
We summarize its time complexity as follows.

\begin{theorem}[Efficient DP for unweighted \knn]
    \label{thm:dp:unweighted}
    The %
    dynamic programming algorithm in \cref{alg:dp:unweighted} for computing the Banzhaf values for the unweighted \knn classifier 
    of all data points runs in $\bigO(nk^2)$ time, where $n$ is the number of data points and $k$ is the number of nearest neighbors. 
\end{theorem}

We demonstrate the algorithm with an example in \cref{ex:dp:unweighted}.

\section{Efficient Monte Carlo estimation of Banzhaf values}
\label{sec:mc}
While the dynamic-programming approach provides exact computation of Banzhaf values, it may become computationally prohibitive for very large datasets due to its quadratic complexity. 
In practice, it is often sufficient to obtain an approximate Banzhaf value.
This motivates us to develop efficient Monte Carlo methods that provide unbiased estimates of Banzhaf values with controllable approximation error.
We introduce two Monte Carlo methods, one by sampling coalitions and the other by sampling permutations.
Both of them are optimized for the \knn value function with significantly stronger statistical efficiency.
The sample complexity can be analyzed using standard techniques, e.g., applying the Chernoff bound; we refer the reader to~\citep{maleki2013bounding,jia2019efficient} for more details.

\subsection{Efficient coalition-based sampling}
\label{sec:mc:coalition}

The most straightforward Monte Carlo approach directly samples from the exponential number of possible coalitions. 
Recall that in \cref{eq:bval-def}, the Banzhaf value of player $i \in N$ is defined as the average marginal contribution of player $i$ to all possible coalitions $S \subseteq N \setminus \{i\}$.
Thus, it can be estimated by uniformly sampling coalitions $S \subseteq N \setminus \{i\}$ and computing the average marginal contribution. 
Specifically, for each player $i \in N$, we sample $m$ coalitions $S_1, S_2, \ldots, S_m$ where each coalition $S_j$ is constructed by including each player in $N \setminus \{i\}$ independently with probability $\frac{1}{2}$. 
Then, the estimate of the Banzhaf value is
\begin{align}
    \bval_i(v) = \frac{1}{m} \sum_{j=1}^{m} \left[ v(S_j \cup \{i\}) - v(S_j) \right]. \label{eq:bval-mc-naive} 
\end{align}

This approach has the advantage of being conceptually straightforward and providing unbiased estimates. 
However, it requires $\bigO(mn)$ coalition evaluations, where each evaluation has complexity $\bigO(n)$ for the \knn value function, resulting in an overall complexity of $\bigO(mn^2)$.

We propose to improve this approach by exploiting the locality property of the \knn value function.
The key idea is to sample one common coalition $S$ where the marginal contribution of all players (data points) in $D$ can be derived simultaneously.
Specifically, we include each data point of $D$ into $S$ independently with probability~$\frac{1}{2}$.
Then, it is sufficient to maintain only the top $k+1$ players in the sorted list of $S$.
To derive the marginal contribution of each player $i$, we remove it from $S$ if it already exists in $S$, or compare it with the $\min\{k, |S|\}$-th closest player in $S$, and decide whether to insert it into $S$.
Equipped with the above optimized operations, the complexity of processing $m$ sampled coalitions for all players can be reduced to $\bigO(mn \log k)$.

\subsection{Efficient permutation-based sampling}
\label{sec:mc:permutation}

We also consider a Monte Carlo method that leverages the permuta\-tion-based formulation of Banzhaf values. 
Permutation-based sampling is mostly well-known for the Shapley values, but it is not directly applicable to the Banzhaf values.
However, we manage to obtain such a formulation by a re-weighting scheme.
Specifically, the Banzhaf values can be expressed as:
\begin{align}
    \bval_i(v) = \frac{n}{2^{n-1}} \sum_{\pi \in \Pi(N)}  \binom{n-1}{|\pi_i|}  \left[ v(\pi_i \cup \{i\}) - v(\pi_i) \right], \label{eq:bval-mc-permutation}
\end{align}
where $\Pi(N)$ denotes the set of all permutations of $N$, and $\pi_i$ represents the set of players that precede player $i$ in $\pi$.
This formulation suggests that we can sample random permutations $\pi$ and compute weighted marginal contributions as the estimate of the Banzhaf value. 
For each sampled permutation $\pi$, we traverse the permutation and compute the marginal contribution of each player when they join the coalition formed by their predecessors. 
Note that each permutation yields marginal contribution estimates for all players simultaneously.
The key insight is that we only need to maintain a sorted list of at most $k$ players in a \knn classifier, and process the traversal along a permutation efficiently within $\bigO(n \log k)$ time.
Since the set of all permutations is of size $n!$, we resort to sampling a subset of $m$ random permutations as an approximation, and the total complexity is $\bigO(mn \log k)$.

\section{Related work}
\label{sec:related}
\paragraph{Data Valuation.}
Data valuation aims to assign importance scores to individual training data points~\citep{hammoudeh2024training,jiang2023opendataval,sim2022data}. 
Simple approaches such as the leave-one-out (LOO) value measure the marginal contribution of a data point to the value function (e.g., model accuracy) when it is removed from the training procedure.
DataShapley~\citep{ghorbani2019data,jia2019towards,wangdata} and its variants such as 
BetaShapley~\citep{kwon2022beta}, 
DataBanzhaf~\citep{wang2023data}, and
least core~\citep{yan2021if}
are all based on the LOO principle, 
but differ in the way the marginal contributions are aggregated.
We discuss several notable options beyond Shapley values below.
\citet{feldman2020neural} simulate the data values by LOO retraining, albeit constrained on a small sample of the training data, 
while DataModels~\citep{ilyas2022datamodels} predict the LOO values by machine-learning models at the expense of exactness.
Another line of popular methods is based on gradients.
TracIn~\citep{pruthi2020estimating} estimates the importance of a training data point by tracing the change in the test loss caused by the data point during the training process.
Variations of influence functions~\citep{koh2017understanding,schioppa2022scaling} have their roots in robust statistics~\citep{cook1982residuals}, 
and offer a gradient-based approximation of the LOO values.

\paragraph{\knn-based Game-theoretic Values.}
Unweighted \knn-Shapley with a soft label was first treated by~\citet{jia2019efficient} and later refined by~\citet{wang2023note}.
The weighted case with a hard label has been addressed by~\citet{wang2024efficient} by dynamic programming with a time complexity of $\bigO(W k^2 n^2)$, which is slower than our \cref{alg:dp:efficient} by a factor of $k$. 
Their key idea is to have the binomial coefficients in \cref{lem:dp:standard} and those in the definition of the Shapley value (see \cref{eq:sval-def-weighted}) cancel out.
However, their technique does not carry over to the \knn-Banzhaf value here, as no binomial coefficient is involved in the definition of the Banzhaf value (\cref{eq:bval-def}).
To the best of our knowledge, we are the first to work on computing Banzhaf values for the \knn problem, 
though Banzhaf values have seen applications in other settings~\citep{wang2023data,karczmarz2022improved}.

\paragraph{Cooperative Game Theory.}
Apart from the Shapley~\citep{shapley1953value} and Ban\-zhaf~\citep{banzhaf1965weighted} values mentioned above, the hard-label \knn-based values are also closely related to majority voting games in cooperative game theory.
It has been shown that weighted majority voting is weakly \np-hard, and there exist dynamic programming algorithms for computing these values (also known as power indices)~\citep{matsui2000survey}.
However, these algorithms are not directly applicable to the \knn-based values, as they do not require a ranking of the players.
Yet, one technique in our unweighted DP algorithm that aggregates the forward and backward functions is inspired by~\citet{uno2012efficient}.
\revise{While our algorithms are tailored to \knn, analogous polynomial-time data-point valuation schemes are uncommon; existing efficient Shapley-value techniques such as TreeSHAP~\citep{lundberg2020local} apply to feature rather than data importance.}

\section{Experiments}
\label{sec:experiments}
We conduct a comprehensive experimental evaluation for \knn classification from the perspectives of computational efficiency and practical applications, including point removal, data selection, and mislabel detection. 
Our main focus is hard-label \knn-Banzhaf valuation in both weighted and unweighted settings, with exact DP and Monte Carlo variants.
We also include hard-label and soft-label Shapley baselines to make the trade-offs explicit, including settings where those baselines are more favorable.
Due to the space limit, we defer missing details of the experiments to \cref{sec:experiments:missing}\citefullpaper.

\paragraph{Datasets.}
\revise{We use 12 classification datasets that vary in size, dimensionality, and number of classes~\citep{kelly2023uci,OpenML2013}.
The datasets are: 
\emph{phoneme}, \emph{wind}, \emph{cpu}, \emph{vehicle}, \emph{pol}, \emph{fraud}, \emph{2dplanes}, \emph{apsfail},
\emph{cifar10}, \emph{mnist},
\emph{creditcard}, and \emph{click}.
Detailed dataset statistics can be found in \cref{tab:dataset-stats}.}
We also use a synthetic dataset %
to evaluate the scalability of the proposed algorithms.

\begin{table}[t]
    \caption{Dataset statistics.}
    \label{tab:dataset-stats}
    \small
    \begin{tabular}{cccc}
    	\toprule
        \textbf{Dataset} & \textbf{\# Data Points} & \textbf{\# Features} & \textbf{\# Classes} \\
        \midrule
        phoneme & 5404 & 5 & 2 \\ %
        wind & 6574 & 14 & 2 \\ %
        cpu & 8192 & 21 & 2 \\ %
        vehicle & 10000 & 100 & 2 \\ 
        pol & 15000 & 48 & 2 \\ %
        fraud & 28000 & 30 & 2 \\ %
        2dplanes & 40768 & 10 & 2 \\ %
        apsfail & 60000 & 170 & 2 \\ %
        cifar10 & 60000 & 3072 & 10 \\ %
        mnist & 70000 & 784 & 10 \\
        creditcard & 284807 & 23 & 2 \\ %
        click & 1000000 & 11 & 2 \\ %
        \bottomrule
    \end{tabular}
\end{table}

\paragraph{Baselines.}
We adopt the following representative baselines:
\begin{itemize}
    \item \texttt{sv} and \texttt{sv-uw}: the DP algorithm for hard-label weighted and unweighted \knn-Shapley \citep{wang2024efficient}.
    \item \texttt{sv-soft}: the analytical solution for soft-label unweighted \knn-Shapley \citep{jia2019efficient}.
    \item \revise{\texttt{beta-mc}: the Beta Shapley values that replace the weights in the Shapley values with a Beta-distributed weights \citep{kwon2022beta}.}
    \item \revise{\texttt{lava}: the Lava values are based on the sensitivity analysis of the class-wise Wasserstein distance between the training and test datasets~\citep{just2023lava}.} 
    \item \texttt{loo}: the classic leave-one-out (LOO) values.
    \item \revise{\texttt{inf}: the influence function which is a gradient-based approximation of the LOO values~\citep{koh2017understanding}.}
    \item \texttt{rand}: the random values.
    \item \texttt{bf}: the brute-force algorithm for \knn-Banzhaf.
\end{itemize}
\revise{We adopt the implementation of \texttt{beta-mc} (with default parameters $\alpha=4,\beta=1$), \texttt{lava} and \texttt{inf} by~\citet{jiang2023opendataval}.}
Algorithms with a suffix of \texttt{-mc} are based on Monte Carlo sampling.

Our algorithms are abbreviated as follows:
\begin{itemize}
    \item \texttt{bv}: the standard DP algorithm for weighted \knn-Banzhaf, described in \cref{sec:alg:standard}.
    \item \texttt{bv-fast}: the improved DP algorithm for weighted \knn-Banzhaf, described in \cref{sec:alg:efficient}.
    \item \texttt{bv-uw}: the near-linear time DP algorithm for unweighted \knn-Banzhaf, described in \cref{sec:alg:efficient-unweighted}.
\end{itemize}

We set the number of nearest neighbors to $k=5$, a uniform weight function, and 5\% of the raw data as the test set for all methods unless otherwise specified.
For weighted \knn, we discretize the weight space of \texttt{bv}, \texttt{bv-fast}, and \texttt{sv} into $7$ bits.
For all experimental results, we report the average over five independent~runs.

\paragraph{Experimental setting.}
All experiments are conducted on a machine with 20 cores of Intel(R) Xeon(R) Gold 6330 CPU @ 2.00GHz and 128 GB RAM, running Ubuntu 20.04.3 LTS with Python 3.8.12.
\rrevise{All runtime measurements use no algorithm-specific parallelism.
We report wall-clock time and use a time budget of 10~hours.}

\begin{figure*}[t]
	\centering
	
    \subcaptionbox{Runtime performance of different algorithms on a synthetic dataset of varying sizes. \label{fig:scalability}}[0.27\textwidth]{
		{\includegraphics[width=0.27\textwidth]{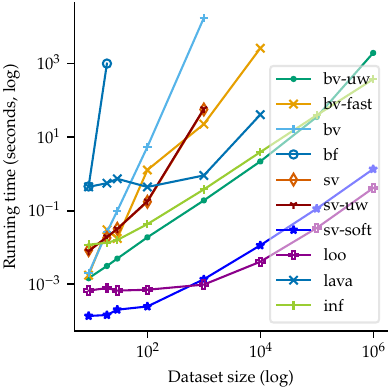}}
	}
    \quad
    \subcaptionbox{The deviation of the Monte Carlo estimates from the unweighted DP estimates as a function of the number of samples.\label{fig:exp:mc-unweighted}}[0.27\textwidth]{
		{\includegraphics[width=0.27\textwidth]{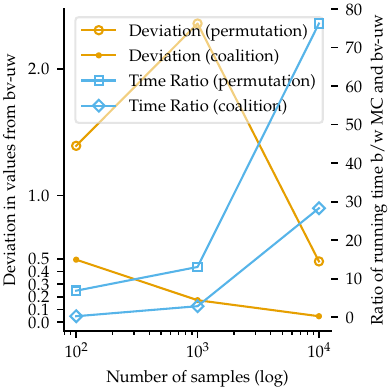}}
	}
    \quad
    \subcaptionbox{The deviation of the Monte Carlo estimates from the weighted DP estimates as a function of the number of samples.\label{fig:exp:mc-weighted}}[0.27\textwidth]{
		{\includegraphics[width=0.27\textwidth]{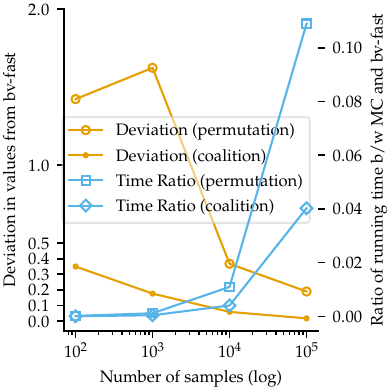}}
	}
	\caption{Computational efficiency of the proposed algorithms.}
\end{figure*}

\subsection{Computational efficiency}
\label{sec:exp:efficiency}

\paragraph{Scalability of the DP algorithms.}
We compare the computational efficiency of our proposed algorithms and other methods by measuring their running time on a synthetic 32-dimensional dataset of varying sizes with one random test point.
We use a uniform weight function.
\rrevise{The results for our DP algorithms and other methods are shown in \cref{fig:scalability}.}

The experimental results confirm that the observed running times of the proposed DP algorithms align closely with our theoretical time complexity analysis presented in \cref{sec:alg}.
The brute-force algorithm cannot finish within 10 hours for a size as small as 30.
The unweighted DP \texttt{bv-uw} indeed has an almost linear time complexity.
Although we fix the parameter $k$, its running time still slightly deviates from the linear one.
This is because of the additional logarithmic overhead of processing huge integers such as those involved in \cref{lem:dp:unweighted-signed} that oversize the machine word size.
\rrevise{Therefore, the nearly linear-time soft-label method \texttt{sv-soft} is more scalable than our \texttt{bv-uw}.}
The improved DP \texttt{bv-fast} is much more scalable than the standard DP \texttt{bv}.
\rrevise{Under the 10-hour budget, the exact hard-label Shapley methods \texttt{sv} and \texttt{sv-uw} are unable to complete runs with $n\geq 10^4$, unlike \texttt{bv-fast} ($10^4$) and \texttt{bv-uw} ($10^6$).}

\paragraph{Performance of the Monte Carlo methods.}
We also compare the performance of the proposed Monte Carlo methods in \cref{sec:mc} with the unweighted and weighted DP algorithms on a dataset with $n=$ 100\,K and 10\,K, respectively.
We define the \emph{deviation} of the Monte Carlo estimates from the exact DP estimates as the maximum absolute difference between the two estimates among all data points, further normalized by the maximum value of the exact DP estimates.
The results are shown in \cref{fig:exp:mc-unweighted} and \cref{fig:exp:mc-weighted} against unweighted \texttt{bv-uw} and weighted \texttt{bv-fast}, respectively.

It is clear that the deviation of the permutation-based Monte Carlo estimate is less stable and converges more slowly than the coalition-based one, given the same number of samples. 
Besides, the former is also more time-consuming due to the maintenance of dynamic top-$k$ neighbors and the additional overhead of computing re-weighting coefficients.
The coalition-based method can provide a speed-up of 25$\times$ over the weighted DP (but not the unweighted DP) if one is willing to accept a deviation that is around $0.02$.
However, to obtain a further smaller deviation, the number of samples required can be large and may not be practical.

\subsection{Point removal}
\label{sec:exp:point-removal}

To examine the ability of the proposed algorithms in discerning data quality, we consider an application that is commonly used in the literature, i.e., point removal.
After obtaining a value for each data point, we gradually remove the data points with the largest values, one after another, and monitor the model performance on the remaining data.
A sharper drop in performance indicates the effectiveness of the valuation.
We use the accuracy of an unweighted \knn classifier over a test set as the performance metric.

\begin{figure*}[t]
	\centering

    \includegraphics[width=0.99\textwidth]{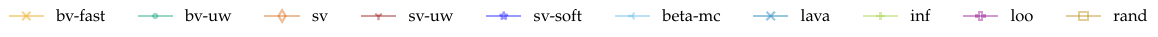}
	
    \subcaptionbox{ \label{fig:rm} Point removal}[0.27\textwidth]{
	{\includegraphics[width=0.27\textwidth]{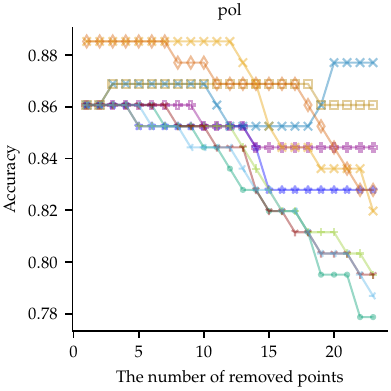}}
	}
    \quad
    \subcaptionbox{ \label{fig:select} Data selection}[0.27\textwidth]{
    {\includegraphics[width=0.27\textwidth]{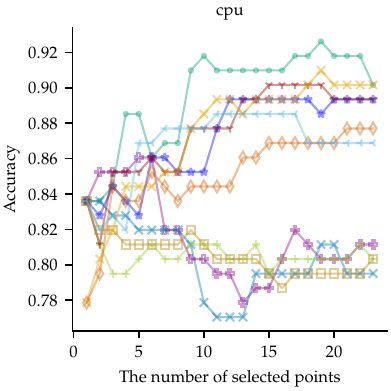}}
    }
    \quad
    \subcaptionbox{ \label{fig:flip} Label flipping}[0.27\textwidth]{
    {\includegraphics[width=0.27\textwidth]{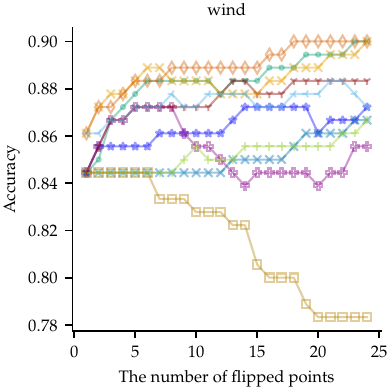}}
    }
    
    \subcaptionbox{ \label{fig:rm2} Point removal}[0.27\textwidth]{
    {\includegraphics[width=0.27\textwidth]{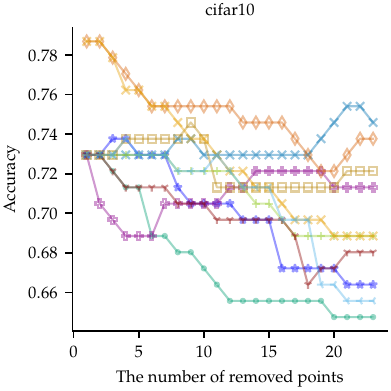}}
    }
    \quad
    \subcaptionbox{ \label{fig:select2} Data selection}[0.27\textwidth]{
    {\includegraphics[width=0.27\textwidth]{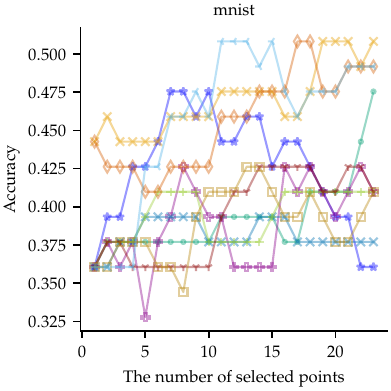}}
    }
    \quad
    \subcaptionbox{ \label{fig:flip2} Label flipping}[0.27\textwidth]{
    {\includegraphics[width=0.27\textwidth]{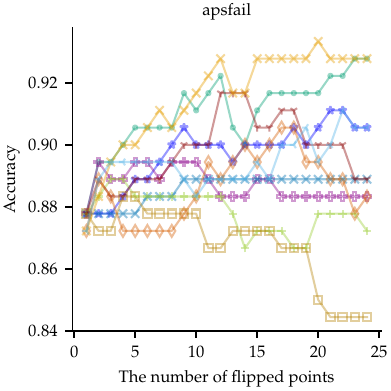}}
    }

	\caption{Point removal results on different datasets: model accuracy as the data points with the largest values are removed (\ref{fig:rm},~\ref{fig:rm2}).
    Data selection results on different datasets: model accuracy as the data points with the largest values are selected (\ref{fig:select},~\ref{fig:select2}).
    The effect of flipping the labels of the data points with the smallest values on the model performance (\ref{fig:flip},~\ref{fig:flip2}).}
\end{figure*}

\revise{
The results are shown in \cref{fig:rm,fig:rm2} (more in \cref{sec:experiments:missing}\citefullpaper).
Among game-theoretic methods,
the hard-label Banzhaf and Shapley values are often the best among many datasets, followed by \texttt{beta-mc} and the soft-label Shapley values \texttt{sv-soft}.
It is worth noting that the weighted \knn methods \texttt{bv-fast} and \texttt{sv} often obtain a higher accuracy than the unweighted ones before removing any data points. %
Baseline \texttt{inf} underperforms, while \texttt{lava} exhibits inconsistent results; 
the model accuracy is expected to drop after removing the most valuable points. 
The accuracy of \texttt{loo} converges much earlier, as it is the simplest measure that only considers the impact of a point on the whole remaining data.
}

\subsection{Data selection}
\label{sec:exp:data-selection}

\revise{
Another popular application of data valuation is data selection, i.e., selecting a subset of data points that are most important to the model performance.
To evaluate this approach, we start with a small number of random data points as the warm-up set, and gradually include more data points with the largest values.
We monitor the model performance as the selected data points are incrementally included in the training set.
The results are shown in \cref{fig:select,fig:select2} (more details in \cref{sec:experiments:missing}\citefullpaper).

Similar to the point-removal experiment, \texttt{loo} and \texttt{rand} are less capable, while \texttt{lava} and \texttt{inf} exhibit inconsistent results.
Among the rest, all game-theoretic methods are fairly comparable to each other, and our proposed algorithms \texttt{bv-fast} and \texttt{bv-uw} are either the best or very close to the best in many datasets.
Our Banzhaf values thus offer a new efficient and robust way to select high-quality data~points.
}

\subsection{Mislabel detection}
\label{sec:exp:mislabel}

Another important application of data valuation is detecting mislabeled training examples, i.e., those data points of a low quality.
\rrevise{We explicitly consider two objectives in this subsection:
(i) detecting injected label noise, and
(ii) identifying low-value points whose correction most improves model performance.}

\rrevise{For objective (i),
we randomly select 5\% of the training data points, and flip their labels to simulate the ground truth of label noise (denoted as $D_{\text{noise}}$). }
After computing the values for the corrupted dataset, we identify the 5\% of data points with the lowest values as our predicted mislabeled points, denoted as $D_{\text{pred}}$.
We measure the detection performance using the $F1$ score between the predicted mislabeled set $D_{\text{pred}}$ and the true mislabeled set $D_{\text{noise}}$, i.e.,
$
F1 = \frac{2 \cdot \text{Precision} \cdot \text{Recall}}{\text{Precision} + \text{Recall}},
$
where 
$
\text{Precision} = \frac{|D_{\text{pred}} \cap D_{\text{noise}}|}{|D_{\text{pred}}|}$, %
and
$\text{Recall} = \frac{|D_{\text{pred}} \cap D_{\text{noise}}|}{|D_{\text{noise}}|}.$
A higher~$F1$ score indicates more accurate detection of mislabeled training examples.

\rrevise{The results show that the soft-label Shapley values (\texttt{sv-soft}) and the influence function (\texttt{inf}) achieve substantially higher $F1$ (and AUC) than the other methods across datasets; detailed per-dataset numbers are provided in \cref{tab:f1,tab:auc-roc}.
Thus, if the sole objective is to recover randomly flipped labels by $F1$ score, these methods are preferable.}

\rrevise{For objective (ii),
we conduct a further experiment to investigate the \emph{influence} of the identified low-value points.}
\rrevise{While soft-label methods detect a larger fraction of the injected mislabeled points, many of the detected points have limited impact on overall model performance.}
The strongest signal for detecting a mislabeled point is its label inconsistency with its neighbors.
Soft-label valuation can capture this signal as soon as some of its neighbors appear in the test set.
In contrast, the hard-label valuation may not consider a noisy point harmful, because it further requires the noisy point to be a game changer for the prediction of its neighbors.
However, when a negative noisy point is surrounded by a majority of positive points, it is unlikely for it to change the prediction.
More importantly, suspected mislabeled points often require further verification by domain experts in practice, which is expensive and time-consuming.
Thus, manually checking mislabeled but irrelevant points is clearly not desirable.

To verify the above analysis, our second experiment flips the labels of the identified mislabeled points (i.e., those with the smallest values), and monitors the model performance on the modified dataset.
The results are shown in \cref{fig:flip,fig:flip2} (more in \cref{sec:experiments:missing}\citefullpaper).
It is evident that the model performance of \texttt{sv-soft} often stays unchanged after flipping the labels of the first few points.
Many of those top points are indeed the mislabeled ones, but they contribute little to the model performance.
\rrevise{In contrast, hard-label valuation like \texttt{bv-fast} can identify the critical low-quality~points, which are not necessarily the random mislabeled ones, and increase the model accuracy significantly.
Therefore, the decision whether to use soft-label or hard-label valuation depends on the application objective: pure noise detection versus identifying \emph{decisive} low-quality points for performance-oriented data cleaning.}

\section{Conclusion}
\label{sec:conclusion}
In this paper, we addressed the computational challenge of computing exact \rrevise{hard-label Banzhaf values for data valuation in $k$-nearest neighbor classifiers}. 
Despite proving that the problem is \sharpp-hard, we successfully exploited the locality properties inherent in \knn classifiers to develop practical and efficient algorithms, including dynamic programming algorithms with significantly improved time complexities: $\bigO(Wkn^2)$ for weighted \knn and $\bigO(nk^2)$ for unweighted \knn.
We also provide efficient Monte Carlo estimation methods, and conduct comprehensive experimental evaluations that demonstrate both computational efficiency and practical effectiveness.

Several promising research directions emerge from this work, including
extending our dynamic-programming framework to other game-theoretic values,
developing hybrid approaches that combine the benefits of hard-label and soft-label valuation methods for more comprehensive data quality assessment, and
exploring the integration of our algorithms with modern deep learning pipelines, in the contexts such as retrieval-augmented generation and embedding-based similarity search.
\revise{It would also be interesting to extend our methods for other machine-learning tasks, 
such as regression and clustering.}

\begin{acks}
This work was supported by 
Guangdong Provincial College Youth Innovative Talent Project (Grant No.\ 2025KQNCX075), 
Natural Science Foundation of Top Talent of SZTU (Grant No.\ GDRC202520),
the ERC Advanced Grant REBOUND (834862),
the Swedish Research Council project ExCLUS (2024-05603),
the Marie Curie Doctoral Training Network ARMADA (101168951), and
the Wallenberg AI, Autonomous Systems and Software Program (WASP) funded by the Knut and Alice Wallenberg Foundation.
\end{acks}

\balance
\bibliographystyle{ACM-Reference-Format}
\bibliography{references}

@misc{kelly2023uci,
    title={The UCI Machine Learning Repository},
    author={Kelly, Markelle and Longjohn, Rachel and Nottingham, Kolby},
    year={2023},
    url={https://archive.ics.uci.edu/}
}

@article{OpenML2013,
  author = {Joaquin Vanschoren and Jan N. van Rijn and Bernd Bischl and Luis Torgo},
  title = {OpenML: networked science in machine learning},
  journal = {SIGKDD Explorations},
  volume = {15},
  number = {2},
  year = {2013},
  pages = {49-60},
  url = {http://doi.acm.org/10.1145/2641190.264119},
  doi = {10.1145/2641190.2641198},
  publisher = {ACM}
}

@article{elkind2016cooperative,
  title={Cooperative game theory},
  author={Elkind, Edith and Rothe, J{\"o}rg},
  journal={Economics and computation: an introduction to algorithmic game theory, computational social choice, and fair division},
  pages={135--193},
  year={2016},
  publisher={Springer}
}

@article{rudin2022interpretable,
  title={Interpretable machine learning: Fundamental principles and 10 grand challenges},
  author={Rudin, Cynthia and Chen, Chaofan and Chen, Zhi and Huang, Haiyang and Semenova, Lesia and Zhong, Chudi},
  journal={Statistic Surveys},
  volume={16},
  pages={1--85},
  year={2022},
  publisher={The American Statistical Association, the Bernoulli Society, the Institute~…}
}

@article{bousquet2002stability,
  title={Stability and generalization},
  author={Bousquet, Olivier and Elisseeff, Andr{\'e}},
  journal={Journal of machine learning research},
  volume={2},
  number={Mar},
  pages={499--526},
  year={2002}
}

@inproceedings{hardt2016train,
  title={Train faster, generalize better: Stability of stochastic gradient descent},
  author={Hardt, Moritz and Recht, Ben and Singer, Yoram},
  booktitle={International conference on machine learning},
  pages={1225--1234},
  year={2016},
  organization={PMLR}
}

@inproceedings{lee2022deduplicating,
  title={Deduplicating Training Data Makes Language Models Better},
  author={Lee, Katherine and Ippolito, Daphne and Nystrom, Andrew and Zhang, Chiyuan and Eck, Douglas and Callison-Burch, Chris and Carlini, Nicholas},
  booktitle={Proceedings of the 60th Annual Meeting of the Association for Computational Linguistics (Volume 1: Long Papers)},
  pages={8424--8445},
  year={2022}
}

@book{gyorfi2002distribution,
  title={A distribution-free theory of nonparametric regression},
  author={Gy{\"o}rfi, L{\'a}szl{\'o} and Kohler, Michael and Krzy{\.z}ak, Adam and Walk, Harro},
  year={2002},
  publisher={Springer}
}

@article{banzhaf1965weighted,
  title={Weighted voting doesn't work: A mathematical analysis},
  author={Banzhaf III, John F},
  journal={Rutgers Law Review},
  volume={19},
  pages={317--343},
  year={1965},
  publisher={HeinOnline}
}

@article{van1998axiomatizations,
  title={Axiomatizations of the normalized Banzhaf value and the Shapley value},
  author={Van den Brink, Rene and Van der Laan, Gerard},
  journal={Social Choice and Welfare},
  volume={15},
  number={4},
  pages={567--582},
  year={1998},
  publisher={Springer}
}

@article{varela2012negotiating,
  title={Negotiating the Lisbon Treaty: Redistribution, Efficiency and Power Indices.},
  author={Varela, Diego and Prado-Dominguez, Javie},
  journal={AUCO Czech Economic Review},
  volume={6},
  number={2},
  year={2012}
}

@inproceedings{wang2023data,
  title = {Data Banzhaf: A Robust Data Valuation Framework for Machine Learning},
  author = {Wang, Jiachen T. and Jia, Ruoxi},
  booktitle = {Proceedings of The 26th International Conference on Artificial Intelligence and Statistics},
  pages = {6388--6421},
  year = {2023},
  publisher = {PMLR}
}

@inproceedings{uno2012efficient,
  title={Efficient computation of power indices for weighted majority games},
  author={Uno, Takeaki},
  booktitle={Algorithms and Computation: 23rd International Symposium, ISAAC 2012, Taipei, Taiwan, December 19-21, 2012. Proceedings 23},
  pages={679--689},
  year={2012},
  organization={Springer}
}

@article{li2023robust,
  title={Robust data valuation with weighted banzhaf values},
  author={Li, Weida and Yu, Yaoliang},
  journal={Advances in Neural Information Processing Systems},
  volume={36},
  pages={60349--60383},
  year={2023}
}

@inproceedings{karczmarz2022improved,
  title={Improved feature importance computation for tree models based on the Banzhaf value},
  author={Karczmarz, Adam and Michalak, Tomasz and Mukherjee, Anish and Sankowski, Piotr and Wygocki, Piotr},
  booktitle={Uncertainty in Artificial Intelligence},
  pages={969--979},
  year={2022},
  organization={PMLR}
}

@incollection{shapley1953value,
    title={A Value for n-Person Games},
    author={Shapley, Lloyd S.},
    booktitle={Contributions to the Theory of Games, Volume {II}},
    year={1953},
    pages={307--318},
    publisher={Princeton University Press},
    address={Princeton, NJ, USA}
}

@article{deng1994complexity,
  author       = {Xiaotie Deng and
                  Christos H. Papadimitriou},
  title        = {On the Complexity of Cooperative Solution Concepts},
  journal      = {Math. Oper. Res.},
  volume       = {19},
  number       = {2},
  pages        = {257--266},
  year         = {1994}
}

@misc{maleki2013bounding,
  title={Bounding the Estimation Error of Sampling-based Shapley Value Approximation}, 
  author={Sasan Maleki and Long Tran-Thanh and Greg Hines and Talal Rahwan and Alex Rogers},
  year={2013},
  eprint={1306.4265},
  archivePrefix={arXiv},
  primaryClass={cs.GT}
}

@article{fatima2008linear,
  author       = {S. Shaheen Fatima and
                  Michael J. Wooldridge and
                  Nicholas R. Jennings},
  title        = {A linear approximation method for the Shapley value},
  journal      = {Artif. Intell.},
  volume       = {172},
  number       = {14},
  pages        = {1673--1699},
  year         = {2008}
}

@article{castro2009polynomial,
  author       = {Javier Castro and
                  Daniel G{\'{o}}mez and
                  Juan Tejada},
  title        = {Polynomial calculation of the Shapley value based on sampling},
  journal      = {Comput. Oper. Res.},
  volume       = {36},
  number       = {5},
  pages        = {1726--1730},
  year         = {2009}
}

@article{bachrach2010approximating,
  title={Approximating power indices: theoretical and empirical analysis},
  author={Bachrach, Yoram and Markakis, Evangelos and Resnick, Ezra and Procaccia, Ariel D and Rosenschein, Jeffrey S and Saberi, Amin},
  journal={Autonomous Agents and Multi-Agent Systems},
  volume={20},
  number={2},
  pages={105--122},
  year={2010},
  publisher={Springer}
}

@article{zhang2023efficient,
  title={Efficient sampling approaches to shapley value approximation},
  author={Zhang, Jiayao and Sun, Qiheng and Liu, Jinfei and Xiong, Li and Pei, Jian and Ren, Kui},
  journal={Proceedings of the ACM on Management of Data},
  volume={1},
  number={1},
  pages={1--24},
  year={2023},
  publisher={ACM New York, NY, USA}
}

@article{mitchell2022sampling,
  title={Sampling permutations for shapley value estimation},
  author={Mitchell, Rory and Cooper, Joshua and Frank, Eibe and Holmes, Geoffrey},
  journal={Journal of Machine Learning Research},
  volume={23},
  number={43},
  pages={1--46},
  year={2022}
}

@inproceedings{ghorbani2019data,
  author       = {Amirata Ghorbani and
                  James Y. Zou},
  title        = {Data Shapley: Equitable Valuation of Data for Machine Learning},
  booktitle    = {Proceedings of the 36th International Conference on Machine Learning},
  pages        = {2242--2251},
  publisher    = {PMLR},
  year         = {2019}
}

@inproceedings{jia2019towards,
  title={Towards efficient data valuation based on the shapley value},
  author={Jia, Ruoxi and Dao, David and Wang, Boxin and Hubis, Frances Ann and Hynes, Nick and G{\"u}rel, Nezihe Merve and Li, Bo and Zhang, Ce and Song, Dawn and Spanos, Costas J},
  booktitle={The 22nd International Conference on Artificial Intelligence and Statistics},
  pages={1167--1176},
  year={2019},
  organization={PMLR}
}

@inproceedings{kwon2022beta,
  title = {Beta Shapley: a Unified and Noise-reduced Data Valuation Framework for Machine Learning},
  author = {Kwon, Yongchan and Zou, James},
  booktitle = {Proceedings of The 25th International Conference on Artificial Intelligence and Statistics},
  pages = {8780--8802},
  year = {2022},
  publisher = {PMLR}
}

@article{yan2021if,
    title={If You Like Shapley Then You'll Love the Core},
    author={Yan, Tom and Procaccia, Ariel D},
    journal={Proceedings of the AAAI Conference on Artificial Intelligence},
    volume={35},
    number={6},
    pages={5751--5759},
    year={2021}
}

@inproceedings{wangdata,
  title={Data Shapley in One Training Run},
  author={Wang, Jiachen T and Mittal, Prateek and Song, Dawn and Jia, Ruoxi},
  booktitle={The Thirteenth International Conference on Learning Representations},
  year = {2025},
  publisher = {OpenReview.net}
}

@article{lundberg2020local,
  title={From local explanations to global understanding with explainable AI for trees},
  author={Lundberg, Scott M and Erion, Gabriel and Chen, Hugh and DeGrave, Alex and Prutkin, Jordan M and Nair, Bala and Katz, Ronit and Himmelfarb, Jonathan and Bansal, Nisha and Lee, Su-In},
  journal={Nature machine intelligence},
  volume={2},
  number={1},
  pages={56--67},
  year={2020},
  publisher={Nature Publishing Group}
}

@article{jia2019efficient,
  author       = {Ruoxi Jia and
                  David Dao and
                  Boxin Wang and
                  Frances Ann Hubis and
                  Nezihe Merve G{\"{u}}rel and
                  Bo Li and
                  Ce Zhang and
                  Costas J. Spanos and
                  Dawn Song},
  title        = {Efficient Task-Specific Data Valuation for Nearest Neighbor Algorithms},
  journal      = {Proc. {VLDB} Endow.},
  volume       = {12},
  number       = {11},
  pages        = {1610--1623},
  year         = {2019}
}

@misc{wang2023note,
  title={A Note on ``Efficient Task-Specific Data Valuation for Nearest Neighbor Algorithms''}, 
  author={Jiachen T. Wang and Ruoxi Jia},
  year={2023},
  eprint={2304.04258},
  archivePrefix={arXiv},
  primaryClass={stat.ML}
}

@inproceedings{wang2024efficient,
  title = {Efficient Data {S}hapley for Weighted Nearest Neighbor Algorithms},
  author = {Wang, Jiachen T. and Mittal, Prateek and Jia, Ruoxi},
  booktitle = {Proceedings of The 27th International Conference on Artificial Intelligence and Statistics},
  pages = {2557--2565},
  year = {2024},
  publisher = {PMLR}
}

@article{matsui2000survey,
  title={A survey of algorithms for calculating power indices of weighted majority games},
  author={Matsui, Tomomi and Matsui, Yasuko},
  journal={Journal of the Operations Research Society of Japan},
  volume={43},
  number={1},
  pages={71--86},
  year={2000},
  publisher={The Operations Research Society of Japan}
}

@book{cook1982residuals,
    title={Residuals and Influence in Regression},
    author={Cook, R. Dennis and Weisberg, Sanford},
    year={1982},
    publisher={Chapman and Hall},
    address={New York, NY, USA}
}

@inproceedings{koh2017understanding,
  author       = {Pang Wei Koh and Percy Liang},
  title        = {Understanding Black-box Predictions via Influence Functions},
  booktitle    = {Proceedings of the 34th International Conference on Machine Learning},
  pages        = {1885--1894},
  publisher    = {PMLR},
  year         = {2017}
}

@article{schioppa2022scaling,
    title={Scaling Up Influence Functions},
    author={Schioppa, Andrea and Zablotskaia, Polina and Vilar, David and Sokolov, Artem},
    journal={Proceedings of the AAAI Conference on Artificial Intelligence},
    volume={36},
    number={8},
    pages={8179--8186},
    year={2022}
}

@article{feldman2020neural,
    title={What Neural Networks Memorize and Why: Discovering the Long Tail via Influence Estimation},
    author={Feldman, Vitaly and Zhang, Chiyuan},
    journal={Advances in Neural Information Processing Systems},
    volume={33},
    pages={2881--2891},
    year={2020}
}

@inproceedings{ilyas2022datamodels,
  title = {Datamodels: Understanding Predictions with Data and Data with Predictions},
  author = {Ilyas, Andrew and Park, Sung Min and Engstrom, Logan and Leclerc, Guillaume and Madry, Aleksander},
  booktitle = {Proceedings of the 39th International Conference on Machine Learning},
  pages = {9525--9587},
  year = {2022},
  publisher = {PMLR}
}

@article{pruthi2020estimating,
    title={Estimating Training Data Influence by Tracing Gradient Descent},
    author={Pruthi, Garima and Liu, Frederick and Kale, Satyen and Sundararajan, Mukund},
    journal={Advances in Neural Information Processing Systems},
    volume={33},
    pages={19920--19930},
    year={2020}
}

@inproceedings{just2023lava,
  title={LAVA: Data Valuation without Pre-Specified Learning Algorithms},
  author={Just, Hoang Anh and Kang, Feiyang and Wang, Tianhao and Zeng, Yi and Ko, Myeongseob and Jin, Ming and Jia, Ruoxi},
  booktitle={The Eleventh International Conference on Learning Representations},
  year={2023},
  organization={OpenReview}
}

@article{hammoudeh2024training,
  author       = {Zayd Hammoudeh and
                  Daniel Lowd},
  title        = {Training data influence analysis and estimation: a survey},
  journal      = {Mach. Learn.},
  volume       = {113},
  number       = {5},
  pages        = {2351--2403},
  year         = {2024}
}

@article{jiang2023opendataval,
  title={OpenDataVal: a Unified Benchmark for Data Valuation},
  author={Kevin Fu Jiang and Weixin Liang and James Y. Zou and Yongchan Kwon},
  journal={Advances in Neural Information Processing Systems},
  volume={36},
  pages={28624--28647},
  year={2023}
}

@inproceedings{sim2022data,
  title     = {Data Valuation in Machine Learning: ``Ingredients'', Strategies, and Open Challenges},
  author    = {Sim, Rachael Hwee Ling and Xu, Xinyi and Low, Bryan Kian Hsiang},
  booktitle = {Proceedings of the Thirty-First International Joint Conference on Artificial Intelligence, {IJCAI-22}},
  publisher = {International Joint Conferences on Artificial Intelligence Organization},
  pages     = {5607--5614},
  year      = {2022}
}

@article{prasad1990np,
	title={NP-completeness of some problems concerning voting games},
	author={Prasad, Kislaya and Kelly, Jerry S},
	journal={International Journal of Game Theory},
	volume={19},
	pages={1--9},
	year={1990},
	publisher={Springer}
}

\clearpage
\appendix

\setcounter{table}{0}
\renewcommand{\thetable}{A\arabic{table}}
\setcounter{figure}{0}
\renewcommand\thefigure{A\arabic{figure}}
\setcounter{algocf}{0}
\renewcommand{\thealgocf}{A\arabic{algocf}}
\setcounter{example}{0}
\renewcommand\theexample{A\arabic{example}}

\section{Full Version of the Specialized DP Algorithm}
\label{sec:dp:unweighted:full}

Full version of the specialized DP algorithm for unweighted \knn in \cref{alg:dp:unweighted:full}.
\begin{algorithm}[t]%
	\DontPrintSemicolon
	\KwIn{Dataset $D$, test point $\ztest$, integer $k$}
	\SetKwComment{tcp}{$\triangleright$\ }{}%
	\SetCommentSty{small}

    Sort $z_1, \ldots, z_n$ by non-decreasing distance to $\ztest$\;
    $W \leftarrow k + 1$\;
    
    \tcp{Compute forward functions $f_i(w,s)$}
    Initialize $f_i(0,0) = 1$ for all $i \in [n]$,  $f_1(w(z_1), 1) = 1$, and others to 0\;
    \For{$i = 2$ \KwTo $n$}{
        \For{$w \in [-W, W]$, $s \in [1, k]$}{
            $f_i(w, s) \leftarrow f_{i-1}(w - w(z_i), s-1) + f_{i-1}(w, s)$\;
        }
    }
    
    \tcp{Compute backward functions $b_i(w,s)$}
    Initialize $b_i(0,0) = 1$ for all $i \in [n+1]$, $b_n(w(z_n), 1) = 1$, and others to 0\;
    \For{$i = n-1$ \KwTo $1$}{
        \For{$w \in [-W, W]$, $s \in [1, k]$}{
            $b_i(w, s) \leftarrow b_{i+1}(w - w(z_i), s-1) + b_{i+1}(w, s)$\;
        }
    }
    
    \tcp{Compute signed backward functions $b_i^+(w,t)$ and $b_i^-(w,t)$}
    Initialize $b_i^+(0,0) = b_i^-(0,0) = 1$ for all $i \in [n+1]$, and others to 0\;
    \For{$t = 1$ \KwTo $k$}{
        \For{$i = n$ \KwTo $1$}{
            \If{$t = 1$}{
                \If{$y_i = \ytest$}{
                    $b_i^+(w(z_i), 1) \leftarrow 2^{n-i} + b_{i+1}^+(w(z_i), 1)$\;
                    $b_i^-(w(z_i), 1) \leftarrow b_{i+1}^-(w(z_i), 1)$\;
                }
                \Else{
                    $b_i^+(w(z_i), 1) \leftarrow b_{i+1}^+(w(z_i), 1)$\;
                    $b_i^-(w(z_i), 1) \leftarrow 2^{n-i} + b_{i+1}^-(w(z_i), 1)$\;
                }
            }
            \Else{
                \For{$w \in [-W, W]$}{
                    $b_i^+(w, t) \leftarrow b_{i+1}^+(w - w(z_i), t-1) + b_{i+1}^+(w, t)$\;
                    $b_i^-(w, t) \leftarrow b_{i+1}^-(w - w(z_i), t-1) + b_{i+1}^-(w, t)$\;
                }
            }
        }
    }

    \tcp{Compute Banzhaf values for all training points}
    \For{$i = 1$ \KwTo $n$}{
        \If{$y_i = \ytest$}{
            $\bval_i(v) \leftarrow \frac{1}{2^{n-1}} (C_- + C_{<k})$ according to \cref{eq:dp:unweighted:bval:positive}\;
        }
        \Else{
            $\bval_i(v) \leftarrow -\frac{1}{2^{n-1}} (C_+ + C_{<k})$ according to \cref{eq:dp:unweighted:bval:negative}\;
        }
    }
    
    \Return{$\{\bval_1(v), \ldots, \bval_n(v)\}$}\;
    
	\caption{Specialized dynamic programming algorithm for unweighted \knn}
	\label{alg:dp:unweighted:full}
\end{algorithm}

\section{Proofs}
\label{sec:proofs}
\printProofs

\section{Missing details of the DP algorithms}
\label{sec:dp:missing}
\begin{example}
    \label{ex:dp:standard}
    Consider a dataset $D = \{z_1, z_2, z_3, z_4\}$ and $k=2$. 
    Let the test point $\ztest$ have label $\ytest = +1$.
    The data points have weights, labels, and ground truth Banzhaf values as follows:
    \begin{center}
    \begin{tabular}{ccccc}
        \toprule
        Point & Distance to $\ztest$ & Label & Weight $w(z_i)$ & Value $\bval_i$ \\
        \midrule
        $z_1$ & 1 & $+1$ & $+3$ & $7/8$ \\
        $z_2$ & 2 & $-1$ & $-2$ & $-1/8$ \\
        $z_3$ & 3 & $+1$ & $+1$ & $1/8$ \\
        $z_4$ & 4 & $-1$ & $-1$ & $-1/8$ \\ 
        \bottomrule
    \end{tabular}
    \end{center}
    
    We compute the Banzhaf value of $z_2$ by constructing $f_2(w, s, m)$ over subsets $S \subseteq D_{\bar{2}} = \{z_1, z_3, z_4\}$.
    
    For $s = 1$ (singletons) we have:
    \begin{align*}
        f_2(3, 1, 1) &= 1 \quad (\text{subset } \{z_1\}),  \\
        f_2(1, 1, 3) &= 1 \quad (\text{subset } \{z_3\}), \text{ and} \\
        f_2(-1, 1, 4) &= 1 \quad (\text{subset } \{z_4\}).
    \end{align*}
    
    For $s = 2$ ($s \le k = 2$) we get:
    \begin{align*}
        f_2(4, 2, 3) &= \sum_{m' < 3: m' \ne 2} f_2(4 - w(z_3), 1, m') \quad\quad (\text{subset } \{z_1, z_3\}) \\
        &= f_2(3, 1, 1) = 1, &\\
        f_2(2, 2, 4) &= \sum_{m' < 4: m' \ne 2} f_2(2 - w(z_4), 1, m') \quad\quad (\text{subset } \{z_1, z_4\}) \\
        &= f_2(3, 1, 1) + f_2(3, 1, 3) = 1 + 0 = 1, \text{ and} &\\
        f_2(0, 2, 4) &= \sum_{m' < 4: m' \ne 2} f_2(0 - w(z_4), 1, m') \quad\quad (\text{subset } \{z_3, z_4\}) \\
        &= f_2(1, 1, 1) + f_2(1, 1, 3) = 0 + 1 = 1. &
    \end{align*}

    For $s = 3$ ($s > k = 2$) we get:
    \begin{align*}
        f_2(4, 3, 3) &= f_2(4, 2, 3) \cdot \binom{4-3}{3-2} = f_2(4, 2, 3) \cdot \binom{1}{1} = 1 \cdot 1 = 1, \text{ and} \\
        f_2(\cdot, 3, 4) &= f_2(\cdot, 2, 4) \cdot \binom{4-4}{3-2} = f_2(\cdot, 2, 4) \cdot \binom{0}{1} = f_2(\cdot, 2, 4) \cdot 0 = 0.
    \end{align*}

    Now we are ready to compute the Banzhaf value. Since $y_2 = -1 \neq \ytest = +1$, we use
    \begin{align*}
        \bval_2(v) = -\frac{1}{2^{3}} 
                    \Biggl( \sum_{\substack{w \in (0, 2]\\ s \in [0,1] \\ m \ne 2}} f_2(w, s, m) + 
                            \sum_{\substack{w \in (0, 2+w(z_m)]\\ s \in [2,3]\\ m > 2}} f_2(w, s, m) \Biggr).
    \end{align*}

    For $C_{<k}$ (subsets of size $< k = 2$), we sum over $w \in (0, 2]$, $s \in [0,1]$ and $m \in \{1, 3, 4\}$:
    \[
        C_{<k} = f_2(1, 1, 3) = 1.
    \]
    
    For $C_{\geq k}$ (subsets of size $\geq k = 2$), we need $w \in (0, 2 + w(z_m)]$ for $m \in \{3, 4\}$ and $s \in [2, 3]$:
    \begin{align*}
        C_{\geq k} &= \sum_{\substack{w \in (0, 2+w(z_3)]\\ s \in [2, 3]}} f_2(w, s, 3) + 
                      \sum_{\substack{w \in (0, 2+w(z_4)]\\ s \in [2, 3]}} f_2(w, s, 4)  \\
        &= 0 + 0 = 0.
    \end{align*}
    
    Therefore, $\bval_2(v) = -\frac{1}{8}(1 + 0) = -\frac{1}{8}$.
    \end{example}

\begin{example}
    \label{ex:dp:unweighted}
    We use the same dataset from \cref{ex:dp:standard} ($n=4$ and $k=2$) but with unweighted \knn models, where the weights become $w(z_i) = +1$ if $y_i = \ytest$ and $w(z_i) = -1$ if $y_i \neq \ytest$.
    Due to space constraints, we skip the computation of the entire DP tables, and only discuss some key entries.
    We compute the Banzhaf value of $z_i = z_3$ ($y_3 = \ytest$) as an example.
    We derive $\bval_i(v) = \frac{1}{8}(C_- + C_{<k})$ as follows:
    \begin{eqnarray*}
        C_- &=& \sum_{\substack{w \in (-k,k)\\ s \in [0,k-1]}} 
            \left( f_{i-1}(w, s) \cdot \sum_{\substack{w'+w \in [-1,0]\\ t = k-s}} b_{i+1}^-(w', t) \right),
    \end{eqnarray*}
    where the non-zero entries include 
    $f_2(0,0) = 1$, $f_2(-1,1) = 1$, ${f_2(1,1)=1}$ and $b_4^-(-1,1) = 1$.
    Therefore,
    \begin{eqnarray*}
        C_- & = & f_2(0,0) \sum_{w'\in[-1,0]} b_4^-(w',2) + f_2(-1,1) \sum_{w'\in[0,1]} b_4^-(w',1)  \\ 
            &   & +~ f_2(1,1) \sum_{w'\in[-2,-1]} b_4^-(w',1) \\
            & = & 1(0) + 1(0) + 1(1) = 1,
    \end{eqnarray*}
    and
    \begin{eqnarray*}
        C_{<k} & = & \sum_{\substack{w \in (-k,k)\\ s \in [0,k-1]}} 
            \left( f_{i-1}(w, s) \cdot B_i(-w, k-1-s) \right),
    \end{eqnarray*}
    where the non-zero backward entries include $b_4(0,0) = 1$, ${b_4(-1,1) = 1}$.
    Therefore,
    \begin{eqnarray*}
        C_{<k} & = & f_2(0,0) B_4(0,1) + f_2(-1,1)B_4(1,0) + f_2(1,1) B_4(-1,0) \\
               & = & 1(1+0) + 1(0) + 1(0) = 1.
    \end{eqnarray*}
    Hence, $\bval_3(v) = \frac{1}{8}(1 + 1) = \frac{1}{4}$.
    
    \end{example}

\subsection{Practical considerations}
\label{sec:dp:practical}

We now discuss some practical considerations, including discretization of the weight space, enumeration of active weights only, and extension to the multi-class setting.

\subsubsection{Discretization of the weight space}
\label{sec:dp:practical:discretization}

As evidenced by \cref{thm:hardness}, no strong polynomial time algorithm is possible for computing the \knn-Banzhaf values for weighted \knn, unless $\p=\np$.
Our DP algorithms in \cref{sec:alg:standard,sec:alg:efficient} are pseudo-polynomial, whose running time depends on the maximum sum of the top-$k$ weights~$W$.
A common practice is to discretize the weight interval $[0,W]$ to avoid the potential exponential blow-up in size.
We follow this approach in our implementation.
Specifically, we discretize the weights into $2^b$ bins that are equally spaced, where $b$ is a user-defined parameter.

\subsubsection{Enumeration of active weights only}
\label{sec:dp:practical:active-weights}

A standard implementation of the DP algorithms enumerates all possible weights in the weight space.
However, in practice, the number of weights that actually exist is usually much smaller.
We call these weights \emph{active weights}.
In our implementation, we keep track of the active weights and only enumerate those during the DP computation.

\subsubsection{Extension to the multi-class setting}
\label{sec:dp:practical:multiclass}

Our framework for binary classification can be naturally extended to the multi-class setting with $|\mathcal{Y}| > 2$ classes using a one-vs-all approach. 
We follow the idea in \citet{wang2024efficient} to decompose the multi-class value function into an average of binary classification subproblems.
Since the Banzhaf value satisfies the linearity property, we can exploit it to solve each binary subproblem independently.

For multi-class classification, we define the value function as the average performance over all binary classification tasks that distinguish the true class $\ytest$ from each incorrect class $c \in \mathcal{Y} \setminus \{\ytest\}$.
Specifically, for a subset $S \subseteq D$ and a test point $\ztest = (\xtest, \ytest)$, we define

\begin{equation}
    v(S \mid \ztest) = \frac{1}{|\mathcal{Y}| - 1} \sum_{c \in \mathcal{Y} \setminus \{\ytest\}} v_c(S \mid \ztest),
    \label{eq:knn-value-function-multiclass}
\end{equation}
where each $v_c(S \mid \ztest)$ is a binary classification value function that distinguishes class $\ytest$ (positive) from class $c$ (negative), with a modified weight function
\[
    w_c(z_i \mid \ztest) = \begin{cases}
    +w(z_i \mid \ztest) & \text{if } y_i = \ytest \\
    -w(z_i \mid \ztest) & \text{if } y_i = c \\
    0 & \text{if } y_i \notin \{\ytest, c\}
    \end{cases}.
\]

By the linearity property of the Banzhaf value, we have
\[
    \bval_i(v \mid \ztest) = \frac{1}{|\mathcal{Y}| - 1} \sum_{c \in \mathcal{Y} \setminus \{\ytest\}} \bval_i(v_c \mid \ztest).
\]
This decomposition allows us to compute the \knn-Banzhaf values in the multi-class setting by solving $|\mathcal{Y}| - 1$ independent binary classification problems for each test point $\ztest$.

\subsubsection{Extension to the regression setting}
\label{sec:dp:practical:regression}

\revise{\citet{wang2023note} introduce the following valuation function for regression:
\[
    v(S \mid \ztest) = \begin{cases}
    -\left(\frac{\sum_{i=1}^{\min\{|S|, k\}} w(z_i(S) \mid \ztest) \, y_i(S)}{\sum_{i=1}^{\min\{|S|, k\}} w(z_i(S) \mid \ztest)} - \ytest \right)^2 & \text{if } |S| > 0 \\
    -\ytest^2 & \text{if } |S| = 0.
    \end{cases}
\]
In other words, it is the negation of the squared loss between the true value and the predicted value by the weighted \knn regressor.

However, it remains a challenging open problem to analytically compute the Banzhaf or Shapley values for the above valuation function.
The main obstacle is the normalization term of weights that is neither decomposable between different subsets (like the soft-label \knn-Shapley~\citep{wang2023note}), nor enumerable (like our dynamic programming approach), due to the division.

We stress that even in the restricted case of unweighted \knn models, the regression setting remains inherently incompatible with our dynamic programming framework, the central contribution of this paper.
The key idea of our framework is to adopt hard-label valuation, which enables one to do efficient counting of pivotal subsets by dynamic programming.
However, hard-label valuation is inherently incompatible with a regressor whose output is a continuous value.
Therefore, it is more plausible in our opinion to stay focused on the dynamic programming framework in this paper for the classification setting.

}

\section{Missing details of the experiments}
\label{sec:experiments:missing}

The dataset statistics can be found in \cref{tab:dataset-stats}.
Image data are fed to a ResNet model to extract features from the penultimate layer.
In order to include the less scalable methods in the comparison, we sample a random subset of several thousand data points.

\begin{figure*}[h]
	\centering

    \includegraphics[width=0.99\textwidth]{pics/rm-legend-ncol10.pdf}
	
    \subcaptionbox{ }[0.22\textwidth]{
        {\includegraphics[width=0.22\textwidth]{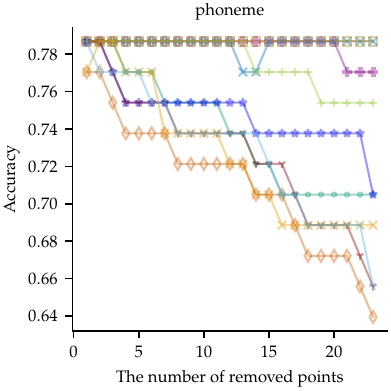}}
    }
    \quad
    \subcaptionbox{ }[0.22\textwidth]{
        {\includegraphics[width=0.22\textwidth]{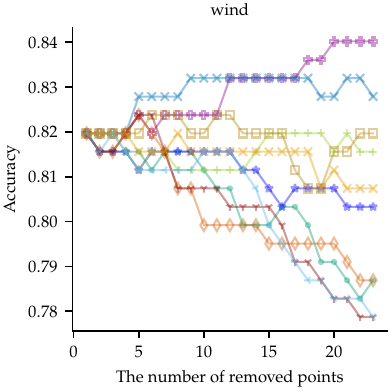}}
    }
    \quad
    \subcaptionbox{ }[0.22\textwidth]{
        {\includegraphics[width=0.22\textwidth]{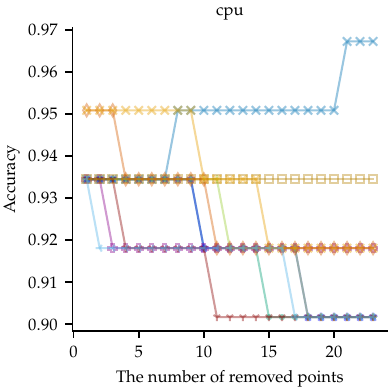}}
    }
    \quad
    \subcaptionbox{ }[0.22\textwidth]{
        {\includegraphics[width=0.22\textwidth]{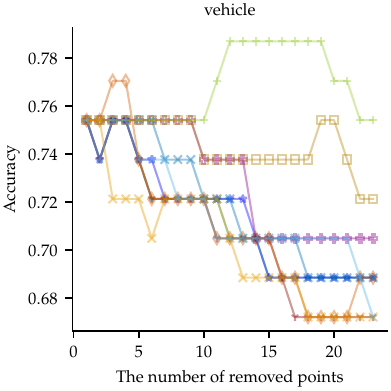}}
    }

    \subcaptionbox{ }[0.22\textwidth]{
        {\includegraphics[width=0.22\textwidth]{pics/rm-pol-knn.pdf}}
    }
    \quad
    \subcaptionbox{ }[0.22\textwidth]{
        {\includegraphics[width=0.22\textwidth]{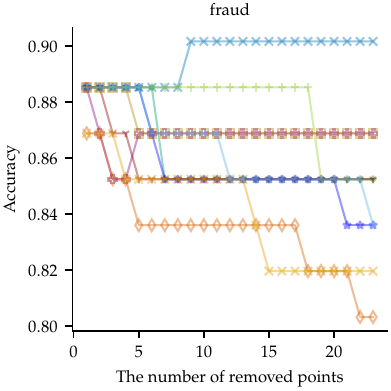}}
    }
    \quad
    \subcaptionbox{ }[0.22\textwidth]{
		{\includegraphics[width=0.22\textwidth]{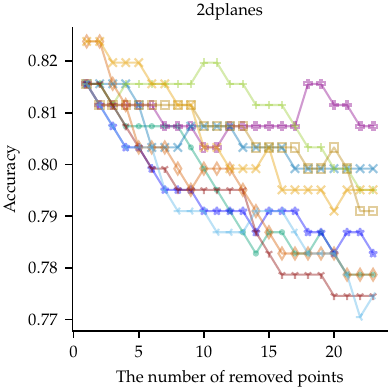}}
	}
    \quad
    \subcaptionbox{ }[0.22\textwidth]{
        {\includegraphics[width=0.22\textwidth]{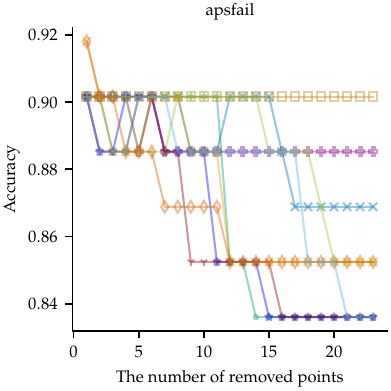}}
    }

    \subcaptionbox{ }[0.22\textwidth]{
        {\includegraphics[width=0.22\textwidth]{pics/rm-cifar10-knn.pdf}}
    }
    \quad
    \subcaptionbox{ }[0.22\textwidth]{
        {\includegraphics[width=0.22\textwidth]{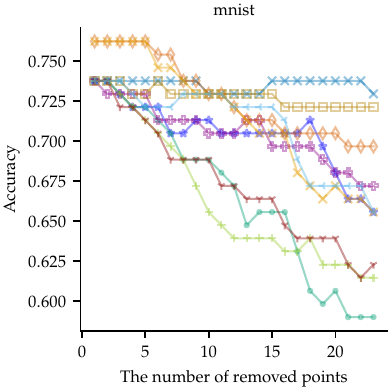}}
    }
    \quad
    \subcaptionbox{ }[0.22\textwidth]{
        {\includegraphics[width=0.22\textwidth]{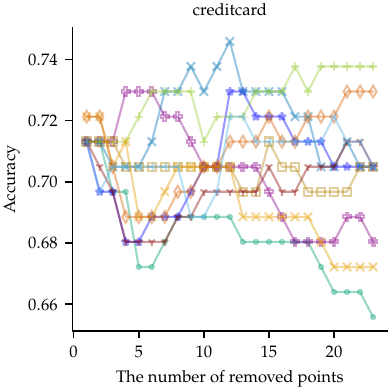}}
    }
    \quad
    \subcaptionbox{ }[0.22\textwidth]{
        {\includegraphics[width=0.22\textwidth]{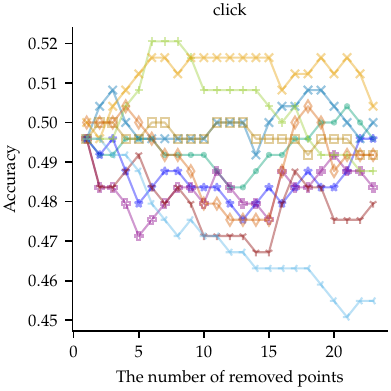}}
    }

    \caption{Point removal results on different datasets: model accuracy as the data points with the largest values are removed.}
    \label{fig:rm-full}
\end{figure*}

\begin{figure*}[h]
	\centering

    \includegraphics[width=0.99\textwidth]{pics/rm-legend-ncol10.pdf}
	
    \subcaptionbox{ }[0.22\textwidth]{
        {\includegraphics[width=0.22\textwidth]{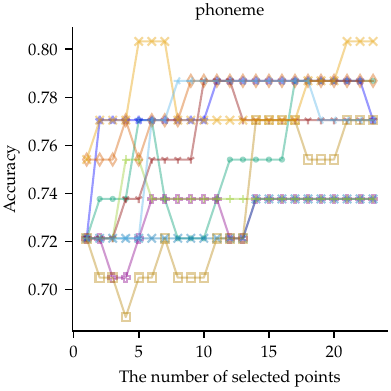}}
    }
    \quad
    \subcaptionbox{ }[0.22\textwidth]{
        {\includegraphics[width=0.22\textwidth]{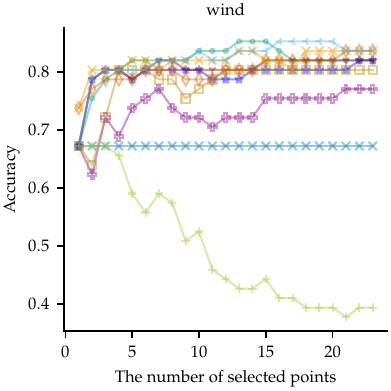}}
    }
    \quad
    \subcaptionbox{ }[0.22\textwidth]{
        {\includegraphics[width=0.22\textwidth]{pics/select-cpu-knn.pdf}}
    }
    \quad
    \subcaptionbox{ }[0.22\textwidth]{
        {\includegraphics[width=0.22\textwidth]{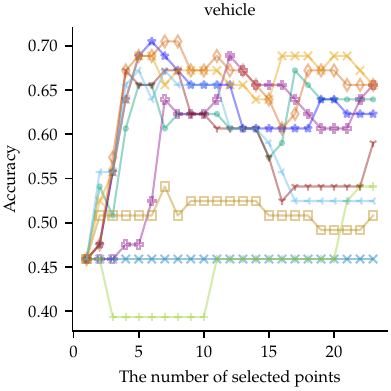}}
    }

    \subcaptionbox{ }[0.22\textwidth]{
        {\includegraphics[width=0.22\textwidth]{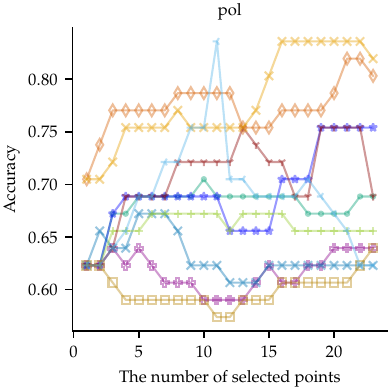}}
    }
    \quad
    \subcaptionbox{ }[0.22\textwidth]{
        {\includegraphics[width=0.22\textwidth]{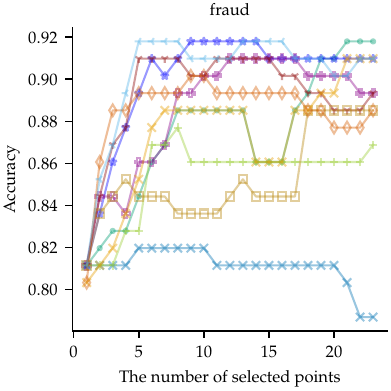}}
    }
    \quad
    \subcaptionbox{ }[0.22\textwidth]{
		{\includegraphics[width=0.22\textwidth]{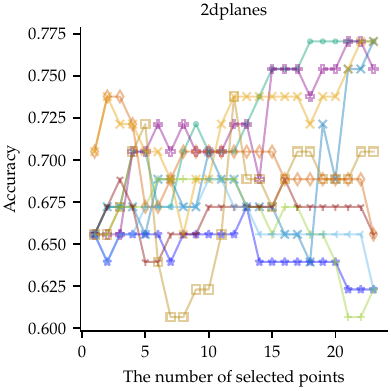}}
	}
    \quad
    \subcaptionbox{ }[0.22\textwidth]{
        {\includegraphics[width=0.22\textwidth]{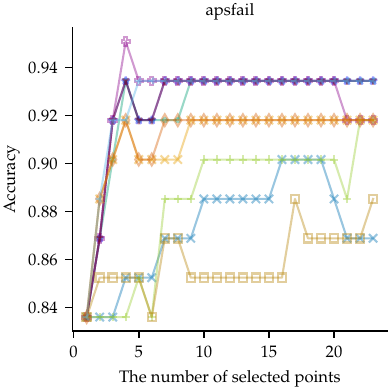}}
    }

    \subcaptionbox{ }[0.22\textwidth]{
        {\includegraphics[width=0.22\textwidth]{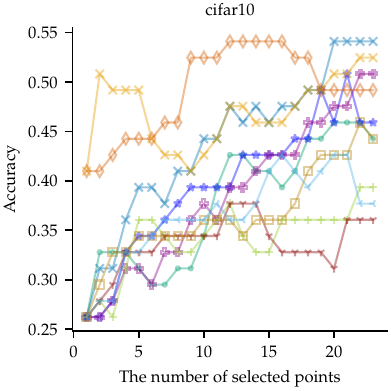}}
    }
    \quad
    \subcaptionbox{ }[0.22\textwidth]{
        {\includegraphics[width=0.22\textwidth]{pics/select-mnist-knn.pdf}}
    }
    \quad
    \subcaptionbox{ }[0.22\textwidth]{
        {\includegraphics[width=0.22\textwidth]{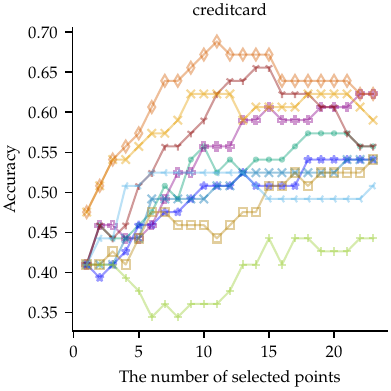}}
    }
    \quad
    \subcaptionbox{ }[0.22\textwidth]{
        {\includegraphics[width=0.22\textwidth]{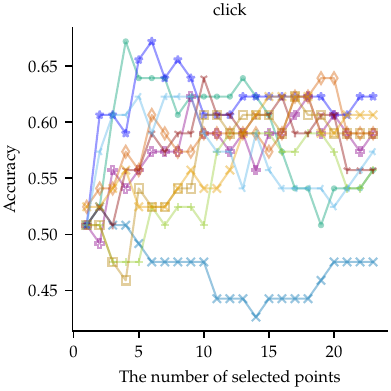}}
    }

    \caption{Data selection results on different datasets: model accuracy as the data points with the largest values are selected.}
    \label{fig:select-full}
\end{figure*}

\begin{figure*}[h]
	\centering
	
    \includegraphics[width=0.99\textwidth]{pics/rm-legend-ncol10.pdf}
	
    \subcaptionbox{ }[0.22\textwidth]{
        {\includegraphics[width=0.22\textwidth]{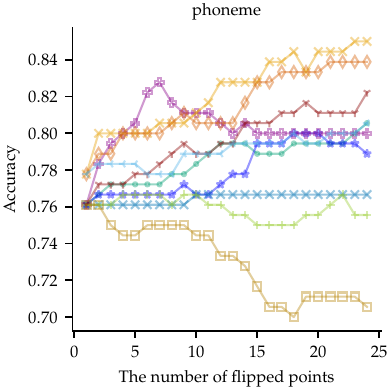}}
    }
    \quad
    \subcaptionbox{ }[0.22\textwidth]{
        {\includegraphics[width=0.22\textwidth]{pics/flip-wind-knn.pdf}}
    }
    \quad
    \subcaptionbox{ }[0.22\textwidth]{
        {\includegraphics[width=0.22\textwidth]{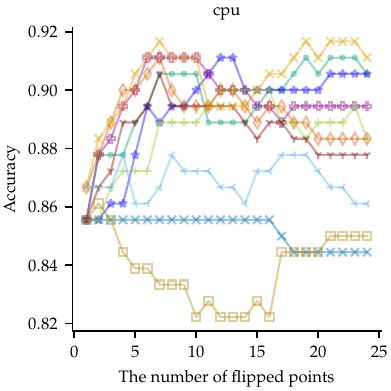}}
    }
    \quad
    \subcaptionbox{ }[0.22\textwidth]{
        {\includegraphics[width=0.22\textwidth]{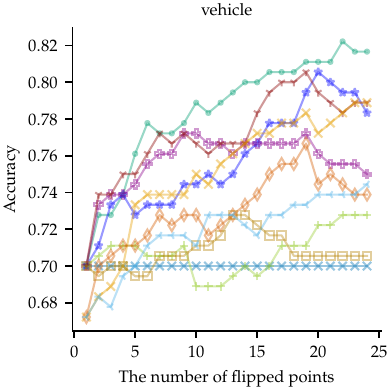}}
    }

    \subcaptionbox{ }[0.22\textwidth]{
        {\includegraphics[width=0.22\textwidth]{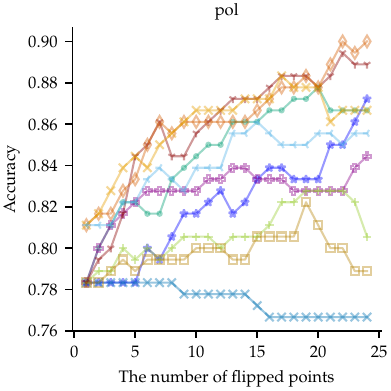}}
    }
    \quad
    \subcaptionbox{ }[0.22\textwidth]{
        {\includegraphics[width=0.22\textwidth]{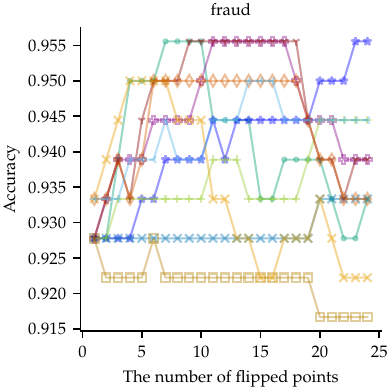}}
    }
    \quad
    \subcaptionbox{ }[0.22\textwidth]{
		{\includegraphics[width=0.22\textwidth]{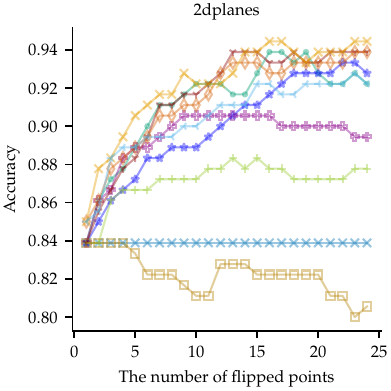}}
	}
    \quad
    \subcaptionbox{ }[0.22\textwidth]{
        {\includegraphics[width=0.22\textwidth]{pics/flip-apsfail-knn.pdf}}
    }

    \subcaptionbox{ }[0.22\textwidth]{
        {\includegraphics[width=0.22\textwidth]{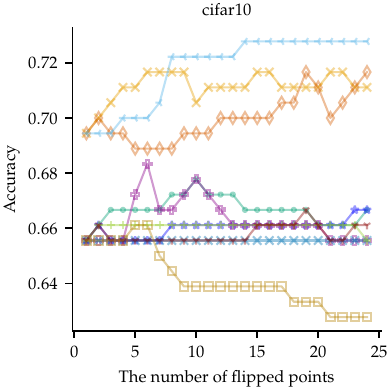}}
    }
    \quad
    \subcaptionbox{ }[0.22\textwidth]{
        {\includegraphics[width=0.22\textwidth]{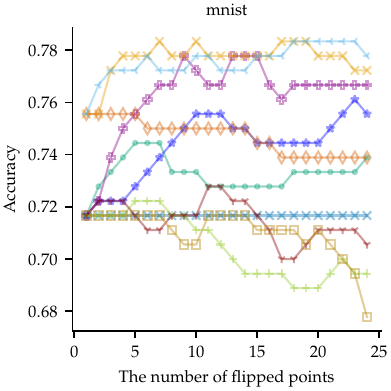}}
    }
    \quad
    \subcaptionbox{ }[0.22\textwidth]{
        {\includegraphics[width=0.22\textwidth]{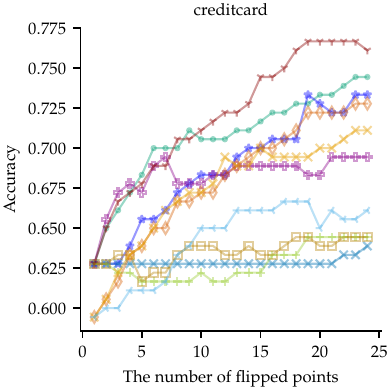}}
    }
    \quad
    \subcaptionbox{ }[0.22\textwidth]{
        {\includegraphics[width=0.22\textwidth]{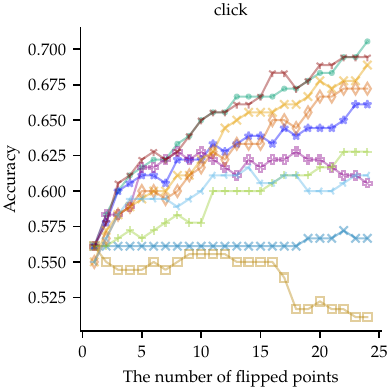}}
    }

	\caption{The effect of flipping the labels of the data points with the smallest values on the model performance.}
    \label{fig:flip-full}
\end{figure*}

See \cref{fig:rm-full}, \cref{fig:select-full}, and \cref{fig:flip-full} for the full experimental results for multiple application tasks.

\subsection{Ablation studies}
\label{sec:exp:ablation}

\begin{figure}[t]
	\centering
	
    \subcaptionbox{The impact of parameter $b$ on the values and running time of \texttt{bv-fast}. \label{fig:ablation:discretization}}[0.22\textwidth]{
		{\includegraphics[width=0.22\textwidth]{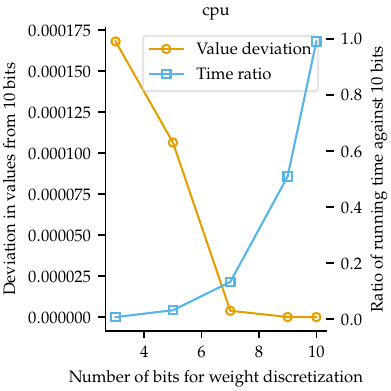}}
	}
    \quad
    \subcaptionbox{The impact of the number of nearest neighbors $k$ on the point removal performance. \label{fig:ablation:k}}[0.22\textwidth]{
		{\includegraphics[width=0.22\textwidth]{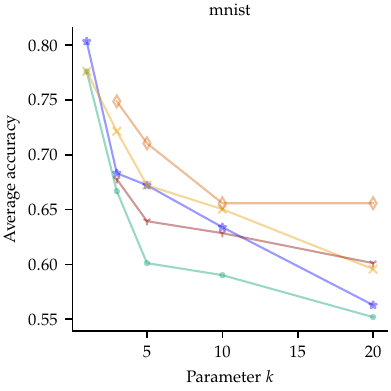}}
	}
    
	\caption{Ablation studies on the proposed algorithms.}
\end{figure}

\revise{
We conduct ablation studies to investigate the contribution of different design decisions of the proposed algorithms.

\paragraph{Discretization of the weight space.}
The time and space complexity of the proposed DP algorithms depend linearly on the size of the maximum sum of the top-$k$ weights $W$.
As discussed in \cref{sec:dp:practical:discretization},
in practice we discretize the weight interval $[0,W]$ into $2^b$ bins that are equally spaced, where $b$ is a user-defined parameter.
We now examine the impact of parameter $b$ on the running time and values of the proposed DP algorithm \texttt{bv-fast}.
The results are shown in \cref{fig:ablation:discretization}.
As expected, the running time increases as the number of bins $b$ increases.
The deviation of the values decreases very quickly as $b$ grows, and plateaus at around $b=7$.
Thus, we recommend using 7 bits for discretization in practice.
}

\revise{
\paragraph{Number of nearest neighbors $k$.}
We also vary the number of nearest neighbors $k$ of the \knn-based algorithms for the task of point removal.
For each $k$, we report the ending accuracy of the model after removing a small number of data points with the largest values.
As shown in \cref{fig:ablation:k}, the ideal number of nearest neighbors is dataset-dependent.
Notably, weighted \knn-based algorithms are less sensitive to the choice of $k$ due to the additional weighting mechanism.
}

\paragraph{Ablation studies on class imbalance}

\begin{figure}[t]
	\centering

    \includegraphics[width=0.22\textwidth]{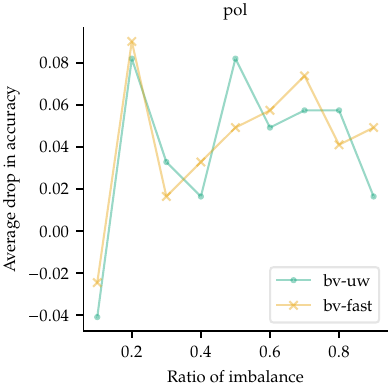}
    \quad
    \includegraphics[width=0.22\textwidth]{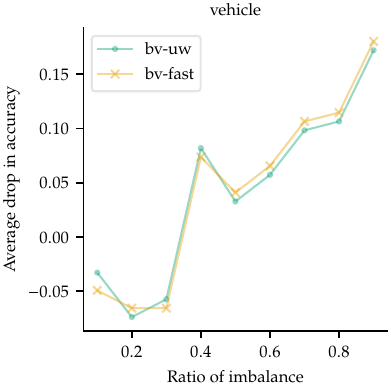}

	\caption{The impact of class imbalance on the performance of the proposed algorithms.}
    \label{fig:ablation:imbalance}
\end{figure}

\revise{
We conduct an ablation study to investigate the impact of class imbalance on the performance of the proposed algorithms.
We vary the ratio of the minority class to the majority class and report the drop in model performance of the point-removal task on the imbalanced dataset.
A larger drop in performance is better.
The results are shown in \cref{fig:ablation:imbalance}.
Overall, the proposed algorithms still perform well and are robust except for extreme cases of class imbalance (e.g., 1:10).
}

\begin{table*}[t]
	\centering
    \caption{$F1$ scores of identifying mislabeled points. The best scores are highlighted in bold, and the second best are underlined.}
	\label{tab:f1}
    \resizebox{\textwidth}{!}{%
    \begin{tabular}{lcccccccccccc}
        \toprule
        & bv-fast & bv-mc & bv-uw & sv & sv-mc & sv-uw & sv-soft & beta-mc & lava & inf & loo & rand \\
        \midrule
        2dplanes & 0.27 $\pm$ 0.13 & 0.24 $\pm$ 0.09 & 0.27 $\pm$ 0.13 & \underline{0.30} $\pm$ 0.09 & \textbf{0.33} $\pm$ 0.04 & \underline{0.30} $\pm$ 0.09 & \textbf{0.33} $\pm$ 0.04 & \textbf{0.33} $\pm$ 0.04 & 0.09 $\pm$ 0.04 & \textbf{0.33} $\pm$ 0.04 & 0.15 $\pm$ 0.04 & 0.06 $\pm$ 0.09 \\
        apsfail & 0.18 $\pm$ 0.00 & 0.21 $\pm$ 0.04 & 0.18 $\pm$ 0.00 & 0.18 $\pm$ 0.00 & 0.18 $\pm$ 0.00 & 0.18 $\pm$ 0.00 & \underline{0.39} $\pm$ 0.04 & 0.18 $\pm$ 0.00 & 0.06 $\pm$ 0.09 & \textbf{0.48} $\pm$ 0.00 & 0.06 $\pm$ 0.09 & 0.00 $\pm$ 0.00 \\
        cifar10 & 0.09 $\pm$ 0.04 & 0.03 $\pm$ 0.04 & 0.09 $\pm$ 0.04 & 0.18 $\pm$ 0.09 & 0.06 $\pm$ 0.00 & 0.18 $\pm$ 0.09 & \textbf{0.48} $\pm$ 0.17 & 0.18 $\pm$ 0.09 & 0.30 $\pm$ 0.26 & \underline{0.39} $\pm$ 0.04 & 0.06 $\pm$ 0.00 & 0.09 $\pm$ 0.04 \\
        click & \textbf{0.09} $\pm$ 0.04 & \underline{0.06} $\pm$ 0.09 & \textbf{0.09} $\pm$ 0.04 & \textbf{0.09} $\pm$ 0.04 & 0.03 $\pm$ 0.04 & \textbf{0.09} $\pm$ 0.04 & 0.03 $\pm$ 0.04 & 0.00 $\pm$ 0.00 & 0.03 $\pm$ 0.04 & 0.00 $\pm$ 0.00 & 0.03 $\pm$ 0.04 & \underline{0.06} $\pm$ 0.09 \\
        cpu & 0.21 $\pm$ 0.04 & 0.18 $\pm$ 0.00 & 0.18 $\pm$ 0.00 & 0.21 $\pm$ 0.04 & 0.39 $\pm$ 0.04 & 0.18 $\pm$ 0.09 & \underline{0.42} $\pm$ 0.09 & 0.39 $\pm$ 0.04 & 0.09 $\pm$ 0.13 & \textbf{0.48} $\pm$ 0.00 & 0.06 $\pm$ 0.00 & 0.03 $\pm$ 0.04 \\
        creditcard & \underline{0.09} $\pm$ 0.04 & \underline{0.09} $\pm$ 0.04 & \underline{0.09} $\pm$ 0.04 & \underline{0.09} $\pm$ 0.04 & \underline{0.09} $\pm$ 0.04 & \underline{0.09} $\pm$ 0.04 & \textbf{0.15} $\pm$ 0.04 & \underline{0.09} $\pm$ 0.04 & 0.03 $\pm$ 0.04 & 0.06 $\pm$ 0.09 & \underline{0.09} $\pm$ 0.04 & 0.06 $\pm$ 0.09 \\
        fraud & 0.18 $\pm$ 0.09 & 0.12 $\pm$ 0.00 & 0.12 $\pm$ 0.00 & 0.24 $\pm$ 0.09 & 0.27 $\pm$ 0.13 & 0.24 $\pm$ 0.09 & \textbf{0.52} $\pm$ 0.13 & 0.39 $\pm$ 0.04 & 0.12 $\pm$ 0.00 & \underline{0.42} $\pm$ 0.17 & 0.06 $\pm$ 0.00 & 0.06 $\pm$ 0.09 \\
        mnist & 0.30 $\pm$ 0.09 & 0.15 $\pm$ 0.13 & 0.21 $\pm$ 0.04 & 0.36 $\pm$ 0.17 & 0.30 $\pm$ 0.26 & 0.27 $\pm$ 0.04 & \underline{0.45} $\pm$ 0.04 & 0.42 $\pm$ 0.17 & 0.18 $\pm$ 0.09 & \textbf{0.48} $\pm$ 0.00 & 0.18 $\pm$ 0.17 & 0.06 $\pm$ 0.09 \\
        phoneme & 0.18 $\pm$ 0.09 & 0.27 $\pm$ 0.04 & 0.12 $\pm$ 0.00 & 0.27 $\pm$ 0.04 & \underline{0.33} $\pm$ 0.04 & 0.18 $\pm$ 0.09 & 0.30 $\pm$ 0.00 & \textbf{0.36} $\pm$ 0.00 & 0.21 $\pm$ 0.04 & 0.27 $\pm$ 0.04 & 0.06 $\pm$ 0.09 & 0.03 $\pm$ 0.04 \\
        pol & 0.09 $\pm$ 0.04 & 0.09 $\pm$ 0.04 & 0.09 $\pm$ 0.04 & 0.18 $\pm$ 0.09 & 0.21 $\pm$ 0.13 & 0.18 $\pm$ 0.09 & \textbf{0.42} $\pm$ 0.00 & 0.18 $\pm$ 0.00 & 0.03 $\pm$ 0.04 & \underline{0.24} $\pm$ 0.00 & 0.18 $\pm$ 0.00 & 0.06 $\pm$ 0.09 \\
        vehicle & 0.12 $\pm$ 0.00 & 0.15 $\pm$ 0.04 & 0.15 $\pm$ 0.04 & \underline{0.18} $\pm$ 0.00 & \underline{0.18} $\pm$ 0.00 & \underline{0.18} $\pm$ 0.00 & \textbf{0.21} $\pm$ 0.04 & \underline{0.18} $\pm$ 0.00 & 0.06 $\pm$ 0.00 & 0.15 $\pm$ 0.04 & 0.12 $\pm$ 0.00 & 0.06 $\pm$ 0.09 \\
        wind & 0.24 $\pm$ 0.00 & 0.24 $\pm$ 0.00 & 0.24 $\pm$ 0.00 & \underline{0.27} $\pm$ 0.04 & \underline{0.27} $\pm$ 0.04 & \underline{0.27} $\pm$ 0.04 & \textbf{0.30} $\pm$ 0.09 & \textbf{0.30} $\pm$ 0.00 & 0.09 $\pm$ 0.04 & 0.24 $\pm$ 0.09 & 0.15 $\pm$ 0.13 & 0.06 $\pm$ 0.09 \\
        \bottomrule
    \end{tabular}%
    }
\end{table*}

\begin{table*}[t]
	\centering
    \caption{Recall scores of identifying mislabeled points. The best scores are highlighted in bold, and the second best are underlined.}
	\label{tab:recall}
    \resizebox{\textwidth}{!}{%
    \begin{tabular}{lcccccccccccc}
        \toprule
        & bv-fast & bv-mc & bv-uw & sv & sv-mc & sv-uw & sv-soft & beta-mc & lava & inf & loo & rand \\
        \midrule
        2dplanes & 0.41 $\pm$ 0.19 & 0.36 $\pm$ 0.13 & 0.41 $\pm$ 0.19 & \underline{0.45} $\pm$ 0.13 & \textbf{0.50} $\pm$ 0.06 & \underline{0.45} $\pm$ 0.13 & \textbf{0.50} $\pm$ 0.06 & \textbf{0.50} $\pm$ 0.06 & 0.14 $\pm$ 0.06 & \textbf{0.50} $\pm$ 0.06 & 0.23 $\pm$ 0.06 & 0.09 $\pm$ 0.13 \\
        apsfail & 0.27 $\pm$ 0.00 & 0.32 $\pm$ 0.06 & 0.27 $\pm$ 0.00 & 0.27 $\pm$ 0.00 & 0.27 $\pm$ 0.00 & 0.27 $\pm$ 0.00 & \underline{0.59} $\pm$ 0.06 & 0.27 $\pm$ 0.00 & 0.09 $\pm$ 0.13 & \textbf{0.73} $\pm$ 0.00 & 0.09 $\pm$ 0.13 & 0.00 $\pm$ 0.00 \\
        cifar10 & 0.14 $\pm$ 0.06 & 0.05 $\pm$ 0.06 & 0.14 $\pm$ 0.06 & 0.27 $\pm$ 0.13 & 0.09 $\pm$ 0.00 & 0.27 $\pm$ 0.13 & \textbf{0.73} $\pm$ 0.26 & 0.27 $\pm$ 0.13 & 0.45 $\pm$ 0.39 & \underline{0.59} $\pm$ 0.06 & 0.09 $\pm$ 0.00 & 0.14 $\pm$ 0.06 \\
        click & \textbf{0.14} $\pm$ 0.06 & \underline{0.09} $\pm$ 0.13 & \textbf{0.14} $\pm$ 0.06 & \textbf{0.14} $\pm$ 0.06 & 0.05 $\pm$ 0.06 & \textbf{0.14} $\pm$ 0.06 & 0.05 $\pm$ 0.06 & 0.00 $\pm$ 0.00 & 0.05 $\pm$ 0.06 & 0.00 $\pm$ 0.00 & 0.05 $\pm$ 0.06 & \underline{0.09} $\pm$ 0.13 \\
        cpu & 0.32 $\pm$ 0.06 & 0.27 $\pm$ 0.00 & 0.27 $\pm$ 0.00 & 0.32 $\pm$ 0.06 & 0.59 $\pm$ 0.06 & 0.27 $\pm$ 0.13 & \underline{0.64} $\pm$ 0.13 & 0.59 $\pm$ 0.06 & 0.14 $\pm$ 0.19 & \textbf{0.73} $\pm$ 0.00 & 0.09 $\pm$ 0.00 & 0.05 $\pm$ 0.06 \\
        creditcard & \underline{0.14} $\pm$ 0.06 & \underline{0.14} $\pm$ 0.06 & \underline{0.14} $\pm$ 0.06 & \underline{0.14} $\pm$ 0.06 & \underline{0.14} $\pm$ 0.06 & \underline{0.14} $\pm$ 0.06 & \textbf{0.23} $\pm$ 0.06 & \underline{0.14} $\pm$ 0.06 & 0.05 $\pm$ 0.06 & 0.09 $\pm$ 0.13 & \underline{0.14} $\pm$ 0.06 & 0.09 $\pm$ 0.13 \\
        fraud & 0.27 $\pm$ 0.13 & 0.18 $\pm$ 0.00 & 0.18 $\pm$ 0.00 & 0.36 $\pm$ 0.13 & 0.41 $\pm$ 0.19 & 0.36 $\pm$ 0.13 & \textbf{0.77} $\pm$ 0.19 & 0.59 $\pm$ 0.06 & 0.18 $\pm$ 0.00 & \underline{0.64} $\pm$ 0.26 & 0.09 $\pm$ 0.00 & 0.09 $\pm$ 0.13 \\
        mnist & 0.45 $\pm$ 0.13 & 0.23 $\pm$ 0.19 & 0.32 $\pm$ 0.06 & 0.55 $\pm$ 0.26 & 0.45 $\pm$ 0.39 & 0.41 $\pm$ 0.06 & \underline{0.68} $\pm$ 0.06 & 0.64 $\pm$ 0.26 & 0.27 $\pm$ 0.13 & \textbf{0.73} $\pm$ 0.00 & 0.27 $\pm$ 0.26 & 0.09 $\pm$ 0.13 \\
        phoneme & 0.27 $\pm$ 0.13 & 0.41 $\pm$ 0.06 & 0.18 $\pm$ 0.00 & 0.41 $\pm$ 0.06 & \underline{0.50} $\pm$ 0.06 & 0.27 $\pm$ 0.13 & 0.45 $\pm$ 0.00 & \textbf{0.55} $\pm$ 0.00 & 0.32 $\pm$ 0.06 & 0.41 $\pm$ 0.06 & 0.09 $\pm$ 0.13 & 0.05 $\pm$ 0.06 \\
        pol & 0.14 $\pm$ 0.06 & 0.14 $\pm$ 0.06 & 0.14 $\pm$ 0.06 & 0.27 $\pm$ 0.13 & 0.32 $\pm$ 0.19 & 0.27 $\pm$ 0.13 & \textbf{0.64} $\pm$ 0.00 & 0.27 $\pm$ 0.00 & 0.05 $\pm$ 0.06 & \underline{0.36} $\pm$ 0.00 & 0.27 $\pm$ 0.00 & 0.09 $\pm$ 0.13 \\
        vehicle & 0.18 $\pm$ 0.00 & 0.23 $\pm$ 0.06 & 0.23 $\pm$ 0.06 & \underline{0.27} $\pm$ 0.00 & \underline{0.27} $\pm$ 0.00 & \underline{0.27} $\pm$ 0.00 & \textbf{0.32} $\pm$ 0.06 & \underline{0.27} $\pm$ 0.00 & 0.09 $\pm$ 0.00 & 0.23 $\pm$ 0.06 & 0.18 $\pm$ 0.00 & 0.09 $\pm$ 0.13 \\
        wind & 0.36 $\pm$ 0.00 & 0.36 $\pm$ 0.00 & 0.36 $\pm$ 0.00 & \underline{0.41} $\pm$ 0.06 & \underline{0.41} $\pm$ 0.06 & \underline{0.41} $\pm$ 0.06 & \textbf{0.45} $\pm$ 0.13 & \textbf{0.45} $\pm$ 0.00 & 0.14 $\pm$ 0.06 & 0.36 $\pm$ 0.13 & 0.23 $\pm$ 0.19 & 0.09 $\pm$ 0.13 \\
        \bottomrule
    \end{tabular}%
    }
\end{table*}

\begin{table*}[t]
	\centering
    \caption{Precision scores of identifying mislabeled points. The best scores are highlighted in bold, and the second best are underlined.}
	\label{tab:precision}
    \resizebox{\textwidth}{!}{%
    \begin{tabular}{lcccccccccccc}
        \toprule
        & bv-fast & bv-mc & bv-uw & sv & sv-mc & sv-uw & sv-soft & beta-mc & lava & inf & loo & rand \\
        \midrule
        2dplanes & 0.20 $\pm$ 0.10 & 0.18 $\pm$ 0.06 & 0.20 $\pm$ 0.10 & \underline{0.23} $\pm$ 0.06 & \textbf{0.25} $\pm$ 0.03 & \underline{0.23} $\pm$ 0.06 & \textbf{0.25} $\pm$ 0.03 & \textbf{0.25} $\pm$ 0.03 & 0.07 $\pm$ 0.03 & \textbf{0.25} $\pm$ 0.03 & 0.11 $\pm$ 0.03 & 0.05 $\pm$ 0.06 \\
        apsfail & 0.14 $\pm$ 0.00 & 0.16 $\pm$ 0.03 & 0.14 $\pm$ 0.00 & 0.14 $\pm$ 0.00 & 0.14 $\pm$ 0.00 & 0.14 $\pm$ 0.00 & \underline{0.30} $\pm$ 0.03 & 0.14 $\pm$ 0.00 & 0.05 $\pm$ 0.06 & \textbf{0.36} $\pm$ 0.00 & 0.05 $\pm$ 0.06 & 0.00 $\pm$ 0.00 \\
        cifar10 & 0.07 $\pm$ 0.03 & 0.02 $\pm$ 0.03 & 0.07 $\pm$ 0.03 & 0.14 $\pm$ 0.06 & 0.05 $\pm$ 0.00 & 0.14 $\pm$ 0.06 & \textbf{0.36} $\pm$ 0.13 & 0.14 $\pm$ 0.06 & 0.23 $\pm$ 0.19 & \underline{0.30} $\pm$ 0.03 & 0.05 $\pm$ 0.00 & 0.07 $\pm$ 0.03 \\
        click & \textbf{0.07} $\pm$ 0.03 & \underline{0.05} $\pm$ 0.06 & \textbf{0.07} $\pm$ 0.03 & \textbf{0.07} $\pm$ 0.03 & 0.02 $\pm$ 0.03 & \textbf{0.07} $\pm$ 0.03 & 0.02 $\pm$ 0.03 & 0.00 $\pm$ 0.00 & 0.02 $\pm$ 0.03 & 0.00 $\pm$ 0.00 & 0.02 $\pm$ 0.03 & \underline{0.05} $\pm$ 0.06 \\
        cpu & 0.16 $\pm$ 0.03 & 0.14 $\pm$ 0.00 & 0.14 $\pm$ 0.00 & 0.16 $\pm$ 0.03 & 0.30 $\pm$ 0.03 & 0.14 $\pm$ 0.06 & \underline{0.32} $\pm$ 0.06 & 0.30 $\pm$ 0.03 & 0.07 $\pm$ 0.10 & \textbf{0.36} $\pm$ 0.00 & 0.05 $\pm$ 0.00 & 0.02 $\pm$ 0.03 \\
        creditcard & \underline{0.07} $\pm$ 0.03 & \underline{0.07} $\pm$ 0.03 & \underline{0.07} $\pm$ 0.03 & \underline{0.07} $\pm$ 0.03 & \underline{0.07} $\pm$ 0.03 & \underline{0.07} $\pm$ 0.03 & \textbf{0.11} $\pm$ 0.03 & \underline{0.07} $\pm$ 0.03 & 0.02 $\pm$ 0.03 & 0.05 $\pm$ 0.06 & \underline{0.07} $\pm$ 0.03 & 0.05 $\pm$ 0.06 \\
        fraud & 0.14 $\pm$ 0.06 & 0.09 $\pm$ 0.00 & 0.09 $\pm$ 0.00 & 0.18 $\pm$ 0.06 & 0.20 $\pm$ 0.10 & 0.18 $\pm$ 0.06 & \textbf{0.39} $\pm$ 0.10 & 0.30 $\pm$ 0.03 & 0.09 $\pm$ 0.00 & \underline{0.32} $\pm$ 0.13 & 0.05 $\pm$ 0.00 & 0.05 $\pm$ 0.06 \\
        mnist & 0.23 $\pm$ 0.06 & 0.11 $\pm$ 0.10 & 0.16 $\pm$ 0.03 & 0.27 $\pm$ 0.13 & 0.23 $\pm$ 0.19 & 0.20 $\pm$ 0.03 & \underline{0.34} $\pm$ 0.03 & 0.32 $\pm$ 0.13 & 0.14 $\pm$ 0.06 & \textbf{0.36} $\pm$ 0.00 & 0.14 $\pm$ 0.13 & 0.05 $\pm$ 0.06 \\
        phoneme & 0.14 $\pm$ 0.06 & 0.20 $\pm$ 0.03 & 0.09 $\pm$ 0.00 & 0.20 $\pm$ 0.03 & \underline{0.25} $\pm$ 0.03 & 0.14 $\pm$ 0.06 & 0.23 $\pm$ 0.00 & \textbf{0.27} $\pm$ 0.00 & 0.16 $\pm$ 0.03 & 0.20 $\pm$ 0.03 & 0.05 $\pm$ 0.06 & 0.02 $\pm$ 0.03 \\
        pol & 0.07 $\pm$ 0.03 & 0.07 $\pm$ 0.03 & 0.07 $\pm$ 0.03 & 0.14 $\pm$ 0.06 & 0.16 $\pm$ 0.10 & 0.14 $\pm$ 0.06 & \textbf{0.32} $\pm$ 0.00 & 0.14 $\pm$ 0.00 & 0.02 $\pm$ 0.03 & \underline{0.18} $\pm$ 0.00 & 0.14 $\pm$ 0.00 & 0.05 $\pm$ 0.06 \\
        vehicle & 0.09 $\pm$ 0.00 & 0.11 $\pm$ 0.03 & 0.11 $\pm$ 0.03 & \underline{0.14} $\pm$ 0.00 & \underline{0.14} $\pm$ 0.00 & \underline{0.14} $\pm$ 0.00 & \textbf{0.16} $\pm$ 0.03 & \underline{0.14} $\pm$ 0.00 & 0.05 $\pm$ 0.00 & 0.11 $\pm$ 0.03 & 0.09 $\pm$ 0.00 & 0.05 $\pm$ 0.06 \\
        wind & 0.18 $\pm$ 0.00 & 0.18 $\pm$ 0.00 & 0.18 $\pm$ 0.00 & \underline{0.20} $\pm$ 0.03 & \underline{0.20} $\pm$ 0.03 & \underline{0.20} $\pm$ 0.03 & \textbf{0.23} $\pm$ 0.06 & \textbf{0.23} $\pm$ 0.00 & 0.07 $\pm$ 0.03 & 0.18 $\pm$ 0.06 & 0.11 $\pm$ 0.10 & 0.05 $\pm$ 0.06 \\
        \bottomrule
    \end{tabular}%
    }
\end{table*}

\begin{table*}[t]
	\centering
    \caption{AUC-ROC scores of identifying mislabeled points. The best scores are highlighted in bold, and the second best are underlined.}
	\label{tab:auc-roc}
    \resizebox{\textwidth}{!}{%
    \begin{tabular}{lcccccccccccc}
        \toprule
        & bv-fast & bv-mc & bv-uw & sv & sv-mc & sv-uw & sv-soft & beta-mc & lava & inf & loo & rand \\
        \midrule
        2dplanes & 0.21 $\pm$ 0.06 & 0.22 $\pm$ 0.06 & 0.21 $\pm$ 0.06 & 0.24 $\pm$ 0.02 & 0.23 $\pm$ 0.01 & 0.24 $\pm$ 0.02 & \textbf{0.27} $\pm$ 0.04 & \underline{0.26} $\pm$ 0.02 & 0.00 $\pm$ 0.00 & 0.23 $\pm$ 0.00 & 0.11 $\pm$ 0.01 & 0.07 $\pm$ 0.10 \\
        apsfail & 0.22 $\pm$ 0.20 & 0.21 $\pm$ 0.18 & 0.11 $\pm$ 0.04 & 0.15 $\pm$ 0.07 & 0.22 $\pm$ 0.16 & 0.15 $\pm$ 0.07 & \underline{0.28} $\pm$ 0.05 & 0.22 $\pm$ 0.16 & 0.05 $\pm$ 0.07 & \textbf{0.43} $\pm$ 0.06 & 0.10 $\pm$ 0.02 & 0.04 $\pm$ 0.06 \\
        cifar10 & 0.09 $\pm$ 0.01 & 0.09 $\pm$ 0.01 & 0.07 $\pm$ 0.01 & 0.19 $\pm$ 0.17 & 0.15 $\pm$ 0.01 & 0.11 $\pm$ 0.05 & \textbf{0.23} $\pm$ 0.11 & 0.21 $\pm$ 0.14 & 0.06 $\pm$ 0.01 & \underline{0.22} $\pm$ 0.10 & 0.07 $\pm$ 0.01 & 0.05 $\pm$ 0.07 \\
        click & 0.04 $\pm$ 0.05 & 0.06 $\pm$ 0.08 & 0.03 $\pm$ 0.05 & 0.06 $\pm$ 0.09 & 0.06 $\pm$ 0.08 & 0.06 $\pm$ 0.09 & 0.04 $\pm$ 0.06 & 0.04 $\pm$ 0.05 & \underline{0.08} $\pm$ 0.05 & \textbf{0.11} $\pm$ 0.05 & 0.01 $\pm$ 0.02 & 0.07 $\pm$ 0.10 \\
        cpu & 0.05 $\pm$ 0.02 & 0.05 $\pm$ 0.02 & 0.05 $\pm$ 0.02 & 0.10 $\pm$ 0.06 & 0.09 $\pm$ 0.03 & 0.10 $\pm$ 0.06 & \textbf{0.31} $\pm$ 0.06 & 0.11 $\pm$ 0.04 & 0.07 $\pm$ 0.09 & \underline{0.27} $\pm$ 0.12 & 0.11 $\pm$ 0.08 & 0.07 $\pm$ 0.01 \\
        creditcard & 0.02 $\pm$ 0.00 & 0.02 $\pm$ 0.01 & 0.02 $\pm$ 0.00 & 0.07 $\pm$ 0.02 & \underline{0.09} $\pm$ 0.03 & 0.07 $\pm$ 0.02 & \textbf{0.11} $\pm$ 0.03 & 0.08 $\pm$ 0.01 & 0.05 $\pm$ 0.03 & 0.05 $\pm$ 0.07 & 0.07 $\pm$ 0.01 & 0.07 $\pm$ 0.10 \\
        fraud & 0.11 $\pm$ 0.01 & 0.12 $\pm$ 0.01 & 0.11 $\pm$ 0.01 & 0.10 $\pm$ 0.06 & 0.11 $\pm$ 0.06 & 0.10 $\pm$ 0.06 & \textbf{0.34} $\pm$ 0.05 & 0.22 $\pm$ 0.05 & 0.00 $\pm$ 0.00 & \underline{0.28} $\pm$ 0.12 & 0.04 $\pm$ 0.06 & 0.05 $\pm$ 0.07 \\
        mnist & 0.11 $\pm$ 0.10 & 0.05 $\pm$ 0.08 & 0.08 $\pm$ 0.06 & 0.15 $\pm$ 0.02 & 0.10 $\pm$ 0.06 & 0.09 $\pm$ 0.07 & \textbf{0.24} $\pm$ 0.11 & 0.18 $\pm$ 0.02 & 0.17 $\pm$ 0.03 & \underline{0.20} $\pm$ 0.12 & 0.01 $\pm$ 0.01 & 0.08 $\pm$ 0.00 \\
        phoneme & 0.15 $\pm$ 0.03 & 0.14 $\pm$ 0.04 & 0.08 $\pm$ 0.07 & 0.18 $\pm$ 0.00 & 0.19 $\pm$ 0.03 & 0.14 $\pm$ 0.06 & \textbf{0.25} $\pm$ 0.05 & \underline{0.23} $\pm$ 0.06 & 0.11 $\pm$ 0.07 & 0.23 $\pm$ 0.06 & 0.11 $\pm$ 0.04 & 0.04 $\pm$ 0.02 \\
        pol & 0.05 $\pm$ 0.01 & 0.06 $\pm$ 0.02 & 0.06 $\pm$ 0.02 & 0.05 $\pm$ 0.01 & 0.05 $\pm$ 0.01 & 0.05 $\pm$ 0.01 & \textbf{0.31} $\pm$ 0.04 & 0.07 $\pm$ 0.07 & 0.05 $\pm$ 0.07 & \underline{0.14} $\pm$ 0.02 & 0.03 $\pm$ 0.04 & 0.07 $\pm$ 0.10 \\
        vehicle & 0.08 $\pm$ 0.00 & 0.10 $\pm$ 0.01 & 0.08 $\pm$ 0.00 & 0.11 $\pm$ 0.03 & 0.11 $\pm$ 0.03 & 0.11 $\pm$ 0.03 & \underline{0.13} $\pm$ 0.01 & 0.09 $\pm$ 0.01 & 0.03 $\pm$ 0.05 & \textbf{0.26} $\pm$ 0.02 & 0.10 $\pm$ 0.03 & 0.07 $\pm$ 0.10 \\
        wind & 0.12 $\pm$ 0.15 & 0.12 $\pm$ 0.13 & 0.12 $\pm$ 0.15 & 0.14 $\pm$ 0.13 & 0.15 $\pm$ 0.13 & 0.14 $\pm$ 0.13 & \underline{0.29} $\pm$ 0.04 & 0.23 $\pm$ 0.05 & 0.07 $\pm$ 0.05 & \textbf{0.31} $\pm$ 0.10 & 0.08 $\pm$ 0.07 & 0.06 $\pm$ 0.00 \\
        \bottomrule
    \end{tabular}%
    }
\end{table*}

\end{document}